\documentclass[10pt,twocolumn,letterpaper]{article}
\usepackage{cvpr}      

\usepackage{tcolorbox}
\usepackage{colortbl}
\usepackage{xcolor}
\usepackage{amsmath}
\usepackage{algorithmic}
\usepackage{algorithm}
\usepackage{bm}
\usepackage{booktabs}
\usepackage{comment}
\usepackage{multirow}
\usepackage{framed}

\usepackage{url}
\makeatletter
\newcommand{\figcaption}[1]{\def\@captype{figure}\caption{#1}}
\newcommand{\tblcaption}[1]{\def\@captype{table}\caption{#1}}
\makeatother

\newcount\K

\def\paragraph#1{\noindent \textbf{#1}}
\usepackage{colortbl,array,xcolor}
\usepackage{wrapfig}

\def\sref#1#2{\hyperref[#1]{\ref*{#1}#2}}
\usepackage{amsthm}
\usepackage{mathtools}

\definecolor{p1}{HTML}{D8C9FF}
\definecolor{p2}{HTML}{CFEFCF}
\definecolor{p3}{HTML}{FFE0A3}
\definecolor{p4}{HTML}{BDE8FF}
\definecolor{p5}{HTML}{D6CEFF}
\definecolor{p6}{HTML}{FFB3B3}
\definecolor{p7}{HTML}{CFEFCF}
\newcommand{\sstrut}{\rule[-0.5ex]{0pt}{2.35ex}}
\newenvironment{toggle}{}{}

\renewcommand{\mid}{\hspace{1pt}|\hspace{1pt}}

\usepackage[whole]{bxcjkjatype} %
\usepackage{xcolor}
\usepackage{colortbl}
\usepackage{multirow}
\definecolor{rowblue}{RGB}{198,234,251}   %
\definecolor{rowgreen}{RGB}{209,239,191}   %
\definecolor{rowgray}{RGB}{245,245,245}   %
\definecolor{highgreen}{RGB}{0,139,69}
\definecolor{commentcolor}{RGB}{237,2,140}   %
\definecolor{lightcyan}{rgb}{0.86,0.93,1}

\newcommand{\midsepremove}{\aboverulesep = -0.16mm \belowrulesep = 0mm}
\midsepremove
\newcommand{\midsepdefault}{\aboverulesep = 0.605mm \belowrulesep = 0.984mm}
\midsepdefault
\definecolor{cvprblue}{rgb}{0.21,0.49,0.74}
\usepackage[pagebackref,breaklinks,colorlinks,allcolors=cvprblue]{hyperref}
\usepackage{placeins}

\def\paperID{2146} %
\def\confName{CVPR}
\def\confYear{2026}

\title{
PowerCLIP: Powerset Alignment for Contrastive Pre-Training
}

\author{Masaki Kawamura$^{1,2}$,
Nakamasa Inoue$^{1,2}$,
Rintaro Yanagi$^{2}$,
Hirokatsu Kataoka$^{2,3}$,
Rio Yokota$^{1,2}$\\
\small{$^{1}$Institute of Science Tokyo, $^{2}$AIST, $^{3}$University of Oxford, VGG} \\
}

\begin{document}
\maketitle

\begin{toggle}
\begin{abstract}
Contrastive vision-language pre-training frameworks such as CLIP have demonstrated impressive zero-shot performance across a range of vision-language tasks.
Recent studies have shown that aligning individual text tokens with specific image patches or regions enhances fine-grained compositional understanding.\hspace{2.5pt}However,\hspace{2.5pt}it remains challenging to capture compositional semantics that span multiple image regions.
To address this limitation, we propose \textbf{PowerCLIP}, a novel
contrastive pre-training framework enhanced by powerset alignment, which exhaustively optimizes region-to-phrase alignments by minimizing the loss defined between powersets of image regions and textual parse trees.
Since the naive powerset construction incurs exponential computational cost due to the combinatorial explosion in the number of region subsets, we introduce efficient non-linear aggregators (NLAs) that reduce complexity from $\mathit{\mathcal{O}(2^{M})}$ to $\mathit{\mathcal{O}(M)}$ with respect to the number of regions $M$, provably approximating the exact loss value with arbitrary precision.
Our extensive experiments demonstrate that PowerCLIP
outperforms state-of-the-art methods
in zero-shot classification and retrieval tasks, underscoring compositionality and robustness of our approach. Code is available at \url{https://github.com/Masakichi210/PowerCLIP}.
\end{abstract}

\begin{figure}
\centering
\includegraphics[width=0.98\linewidth]{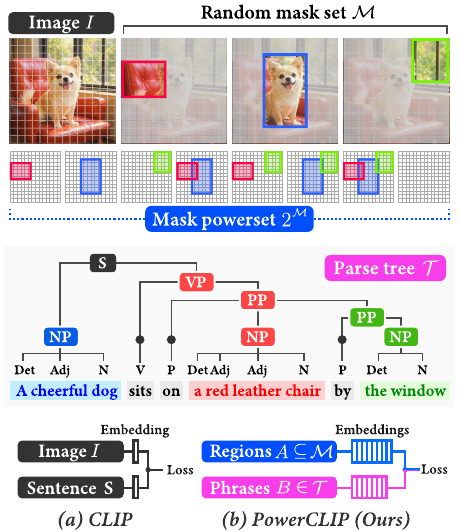}
\vspace{-5pt}
\caption{\textbf{Overview of PowerCLIP.}\hspace{2pt}(a) CLIP aligns images and sentences globally.\hspace{2pt}(b) PowerCLIP explores all combinations of
image regions (\textit{i.e.}\hspace{-0.5pt},\hspace{1pt}powerset) and aligns them with textual phrases.
}
\label{fig:top}
\vspace{-12pt}
\end{figure}

\begin{figure*}
\centering
\includegraphics[width=\linewidth]{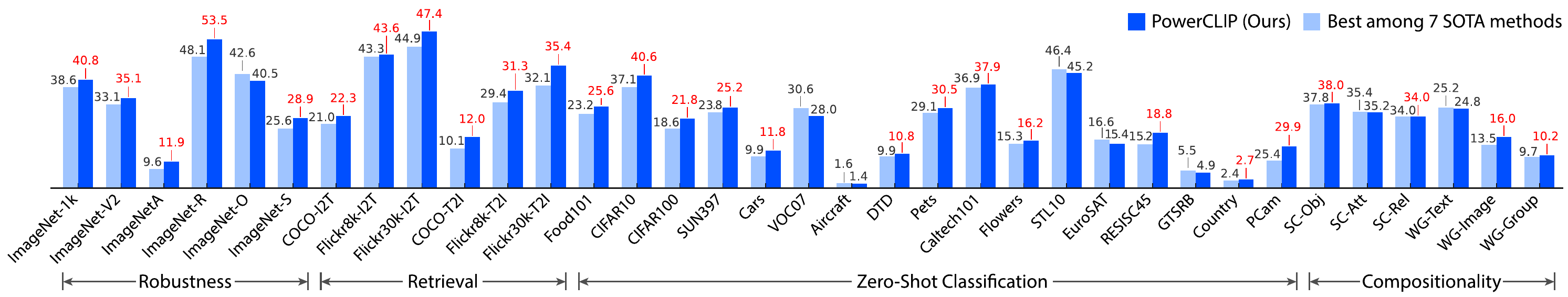}
\vspace{-18pt}
\caption{Performance comparison between PowerCLIP and the best-performing method among seven state-of-the-art approaches (CLIP, FLIP, A-CLIP, E-CLIP, C-PGS, FILIP, and SPARC). Performance improvements are highlighted in red.}
\label{fig:performance}
\vspace{-14pt}
\end{figure*}

\section{Introduction}
Large-scale contrastive pre-training has established a robust foundation for vision-language understanding.\hspace{2pt}A prominent example is CLIP~\cite{clip}, which aligns visual and textual embeddings within a shared semantic space by minimizing the image-text contrastive loss. To improve robustness and compositionality, recent studies have explored sophisticated local and global alignment techniques~\cite{Yang2023a-clip,  Yao2022FILIP,  Patel2024TripletCLIP, Wei2024e-clip, Asokan2025FineLIP, Choi2025GOAL, Pei2025CLIPPGS}.
Local alignment approaches, such as SPARC~\cite{Bica2024SPARC} and FineLIP~\cite{Asokan2025FineLIP}, explicitly match textual tokens with corresponding visual patches, facilitating fine-grained correspondences.
Global alignment approaches, such as A-CLIP~\cite{Yang2023a-clip} and CLIP-PGS~\cite{Pei2025CLIPPGS}, emphasize semantically informative image regions by applying global masks to visual patches.
Though effective, both paradigms operate under single-region or masked-region objectives, limiting their ability to capture compositions among multiple visual entities.
Motivated by this limitation, we propose \textbf{PowerCLIP}, a novel local-to-global alignment framework, which exhaustively aligns image regions with structured textual phrases in a combinatorial manner.

The core idea behind PowerCLIP lies in the \textit{powerset alignment} strategy, which systematically explores all possible subsets of image regions (\textit{i.e.}, the powerset of image regions) and aligns them with phrase structures extracted from textual parse trees.
Specifically, since pre-training begins from scratch, we first generate a set of random region masks $\mathcal{M}$ for each image and then define a contrastive objective between the powerset $2^{\mathcal{M}}$ and the textual parse tree $\mathcal{T}$, as illustrated in Figure~\ref{fig:top}.
This approach significantly enhances the compositionality and robustness due to the exhaustive exploration of local-to-global alignments.

Moreover, since powerset alignment inherently introduces exponential computational complexity, we develop theoretically grounded approximations using \textbf{Non-Linear Aggregators (NLAs)} that reduce the complexity to linear in terms of the number of region masks.
Through extensive experimentation, we demonstrate that PowerCLIP achieves state-of-the-art performance across 22 out of 28 diverse benchmarks, including classification, image-text retrieval, robustness, and compositionality evaluations, as shown in Figure~\ref{fig:performance}.
Our key contributions are summarized as follows:
\begin{itemize}
\item We propose PowerCLIP, a novel contrastive pre-training framework leveraging powerset alignment between image regions and textual phrases.
\item We develop NLAs that derive computationally tractable approximations for powerset alignment, reducing complexity from exponential to linear. We prove that NLAs approximate the exact loss value with arbitrary precision under mild assumptions (Theorems 1 and 2).
\item We demonstrate that PowerCLIP attains state-of-the-art performance across diverse zero-shot benchmarks, improving compositional reasoning and robustness.
\end{itemize}

\section{Related Work}

\paragraph{Contrastive \hspace{-1pt}Pre-training.}
Image-text contrastive learning, pioneered by methods such as CLIP~\cite{clip} and ALIGN~\cite{Jia2021ALIGN},
has become a cornerstone in large-scale vision-language pre-training~\cite{Mu2022SLIP,Yu2022CoCa,Sun2023EVACLIP,Zhai2023SigLIP,Xu2024MetaCLIP,Tschannen2025SigLIP2,Chuang2025MetaCLIP2}.
Meanwhile, several studies highlight its limitations, particularly regarding compositionality and robustness, due to inherent difficulties in embedding complex visual and textual structures into a single shared semantic space~\cite{Hsieh2023SugarCrepe, Dumpala2024SUGARCREPEpp,Thrush2022Winoground, Kang2025DCSM}.
Recent efforts to address these limitations have focused on improving alignment from visual, textual, and multimodal perspectives.

\paragraph{Visual\hspace{2pt}Masking\hspace{2pt}Approaches.}
Inspired by masked image modeling~\cite{He2022MAE,Xie2022SimMIM,Wei2022MaskFeat}, visual masking approaches have significantly enhanced global image-text alignment.
For example, FLIP~\cite{Li2023flip} applies random masks for efficient and robust training. 
MaskCLIP~\cite{Dong2023MaskCLIPDistil} incorporates masking mechanism into self-distillation.
Subsequent approaches have focused more on structured and targeted masking.
A-CLIP~\cite{Yang2023a-clip} emphasizes informative image regions through attentive masking.
E-CLIP~\cite{Wei2024e-clip} employs cluster masking to better capture visual structures.
CLIP-PGS~\cite{Pei2025CLIPPGS} proposes gradual masking with the patch generation-to-selection mechanism.
In contrast to these approaches, PowerCLIP performs local-to-global alignment by exploring combinations of region masks and aligns them with textual structures, enhancing compositionality and robustness.

\paragraph{Textual\hspace{2pt}Approaches.}
From the textual perspective, several methods for textual augmentation and refinement have been proposed. For instance, VeCLIP~\cite{Lai2024VeCLIP} enriches textual descriptions, while LaCLIP~\cite{Fan2023LaCLIP} rewrites them to better align with visual semantics. NegationCLIP~\cite{Park2025NegationCLIP} introduces negation terms into textual descriptions to provide richer contrasts. Synthetic approaches have also shown promising effectiveness~\cite{Yuksekgonul2023NegCLIP, Wei2025HQCLIP, Patel2024TripletCLIP}.
TripletCLIP~\cite{Patel2024TripletCLIP} demonstrates that a triplet contrastive loss with hard negative samples improves compositionality.
Although we retain the original text during pre-training for a fair comparison with other types of approaches, these studies motivate us to design a triplet margin loss to enhance compositionality.

\paragraph{Multimodal\hspace{2pt}Approaches.}\hspace{3pt}Multimodal approaches primarily target fine-grained alignment between textual tokens and visual patches.
Prominent examples include FILIP~\cite{Yao2022FILIP}, which performs token-level alignment via cross-modal late interaction; SPARC~\cite{Bica2024SPARC}, which employs sparse cross-modal alignment; and LAPS~\cite{Fu2024LAPS}, which aligns patches and words by identifying redundant visual regions. In fine-tuning scenarios, several methods have addressed alignment with longer textual descriptions via fine-tuning or incremental training, such as LongCLIP~\cite{Zhang2024LongCLIP}, FixCLIP~\cite{Wang2025FixCLIP}, GOAL~\cite{Choi2025GOAL}, and FineLIP~\cite{Asokan2025FineLIP}.
Additionally, word-to-region correspondences have been explored for downstream tasks via fine-tuning and adaptation for object detection~\cite{Zhou2022MaskCLIP_Extra,Zhong2022RegionCLIP,Li2025DenseVLM,Sun2024AlphaCLIP,li2021GLIP} and semantic segmentation~\cite{Jing2024FineCLIP,Jose2025DINOv2MeetsText,Li2025MaskAdapter,Choi2025PartCATSeg,Peng2025HCLIP,Zhang2025CorrCLIP,Ge2025CRTNet,Duan2025DIHCLIP,Xie2025FG-CLIP,Mulhoti2023PACL}.
However, capturing compositional semantics across multiple image regions remains challenging.
We focus on pre-training scenarios and address compositionality by aligning combinations of image regions with textual parse trees, facilitating effective local-to-global alignment.

\begin{figure*}
\centering
\includegraphics[width=\linewidth]{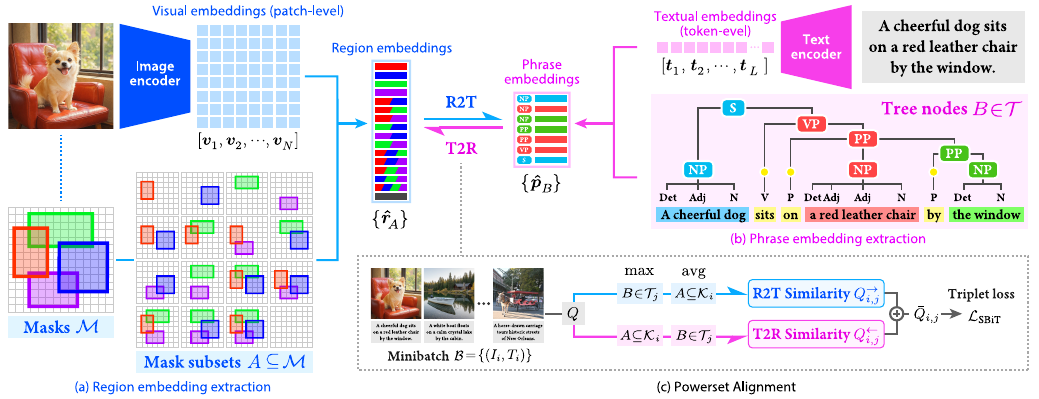}
\vspace{-16pt}
\caption{\textbf{Overview of the powerset alignment strategy for PowerCLIP.}
(a) Region embeddings are extracted for each subset $A$ of region masks in $\mathcal{M}$.
(b) Phrase embeddings are extracted for each node $B$ in the parse tree $\mathcal{T}$.
(c) Powerset alignment minimizes the triplet loss defined based on the bidirectional similarity: region-set-to-tree (R2T) and vice versa (T2R).
}
\label{fig:overview}
\vspace{-12pt}
\end{figure*}

\section{Method}

This section introduces \textbf{PowerCLIP}, a novel contrastive pre-training framework for local-to-global alignment. The core idea behind PowerCLIP is powerset alignment, which exhaustively explores combinatorial correspondences between image regions and textual phrases, improving compositionality and robustness.

\subsection{Overview}

\paragraph{Problem Setting and Notation.} 
We study image-text contrastive pre-training, where the training dataset consists of images paired with their corresponding textual descriptions.
To achieve fine-grained alignment, we adopt Transformer-based encoders for both the image and text modalities to extract patch-level and token-level embeddings, respectively.
We denote visual embeddings extracted from an image $I$ by $[\bm{v}_{1}, \bm{v}_{2}, \cdots\hspace{-1pt}, \bm{v}_{N}]\!\in\!\mathbb{R}^{D \times N}$, and textual embeddings from a text description $T$ by $[\bm{t}_{1}, \bm{t}_{2}, \cdots\hspace{-1pt}, \bm{t}_{L}]\hspace{-1pt}\!\in\!\hspace{-0.5pt}\mathbb{R}^{D \times L}$, where $N$ is the number of image patches, $L$ is the length of the token sequence, and $D$ is the shared feature dimension.

\paragraph{Architecture.}
Figure~\ref{fig:overview} shows the architectural overview of PowerCLIP, which aligns powersets of image regions and textual parse trees in three primary steps.
First, for each training image $I$, a set of region masks $\mathcal{M}$ is generated on a patch grid either randomly or via a segmentation model. Here, region embeddings corresponding to all subsets of masks $A \subseteq \mathcal{M}$ are extracted as candidates to be matched with textual phrases, as shown in Figure~\sref{fig:overview}{(a)}.
Second, phrase embeddings are extracted from each textual description $T$ by identifying phrases using a parse tree as shown in Figure~\sref{fig:overview}{(b)}.
Finally, powerset alignment is performed by minimizing the triplet loss defined with similarities in two directions: region-set-to-tree (R2T) and vice versa (T2R).
Compared with conventional alignment methods (\textit{e.g.}, SPARC ~\cite{Bica2024SPARC}), our approach considers candidate matches more exhaustively over possible combinations of image regions and textual phrases, enhancing compositionality in image-text contrastive learning.
Below, we describe details for region embedding extraction (\S\ref{sec:region_embedding}), phrase embedding extraction (\S\ref{sec:phrase_embedding}), and powerset alignment (\S\ref{sec:powerset_alignment}).

\subsection{Region Embedding Extraction}
\label{sec:region_embedding}
\paragraph{Region Masks.}
For each image $I$, we randomly generate $M\!\in\!\mathbb{N}$ bounding boxes on the patch grid by uniformly sampling their centers, heights, and widths.
These bounding boxes define the set of region masks $\mathcal{M}\!=\!\{R_{m}\}_{m=1}^{M}$, where each $R_{m}\!\in\!\{0,1\}^{N}$ is a binary mask over patches.
Optionally, this step can utilize segmentation models such as SAM~\cite{ravi2024sam2} instead of random sampling.

\paragraph{Region Embeddings.}
To allow comprehensive matching with textual structures, we construct the powerset of the set of region masks: $2^{\mathcal{M}}\!=\!\{ A\!\subseteq\!\mathcal{M} \}$, where each subset $A$ corresponds to a combination of region masks.
We then define the region embeddings $\bm{r}_{A}$ for each $A$ by
\begin{align}
\bm{r}_{A}
=
\sum_{R_{m} \in A}
\phi(I \mid R_{m})
\end{align}
where $\phi$ is a function to encode the image $I$ given an individual region mask $R_{m}$.
For computational efficiency, we apply masks to visual embeddings obtained from the entire image rather than encoding each image region independently.
Specifically, we define $\phi$ by
{
\setlength{\abovedisplayskip}{2pt}
\setlength{\belowdisplayskip}{4pt}
\begin{align}
\label{eq:phi}
\phi(I \mid R_{m})
=
\frac{\bm{r}_{m}}{\| \bm{r}_{m} \|_{2}},
\quad
\bm{r}_{m} = \sum_{n=1}^{N} R_{mn} \bm{v}_{n},
\end{align}
where embeddings are L2 normalized.
}

\subsection{Phrase Embedding Extraction}
\label{sec:phrase_embedding}

\paragraph{Parse Trees.}
For each text description $T$, we generate a constituency parse tree $\mathcal{T}$ by applying a syntactic parser.
Each node $B\!\in\!\mathcal{T}$ corresponds to a sentence-level or phrase-level constituent, \textit{e.g.}, Noun Phrase (NP), Verb Phrase (VP), Prepositional Phrase (PP), or Sentence (S).

\paragraph{Token Masks.}
Analogous to the region masks for the visual modality, we represent each leaf node by a token mask $P_{m'}\!\in\!\{0,1\}^{L}$, where $m'$ indexes the leaf nodes.
For example, given the description
\textit{``\colorbox{p1}{\sstrut \hspace{0.2pt}a\hspace{0.2pt}}\colorbox{p2}{\sstrut ~\hspace{1pt}dog}\colorbox{p3}{\sstrut ~\hspace{1pt}sitting}\colorbox{p4}{\sstrut ~\hspace{1pt}on}\colorbox{p5}{\sstrut ~\hspace{1pt}a}\colorbox{p6}{\sstrut ~\hspace{1pt}red}\colorbox{p7}{\sstrut ~\hspace{1pt}chair},''} the noun phrase \textit{``a dog''} is represented by a mask assigning ones to tokens \textit{\colorbox{p1}{\sstrut \hspace{0.2pt}a\hspace{0.2pt}}} and \textit{\colorbox{p2}{\sstrut ~\hspace{1pt}dog}} and zeros elsewhere.
Consequently, each non-leaf node $B$ is represented by a set of token masks corresponding to its leaf nodes.

\paragraph{Phrase Embeddings.}
We define the phrase embeddings $\bm{p}_{B}$ for each node $B \in \mathcal{T}$ by
{
\setlength{\abovedisplayskip}{4pt}
\setlength{\belowdisplayskip}{4pt}
\begin{align}
\bm{p}_{B} = \sum_{P_{m'} \in B} \psi(T \mid P_{m'}) 
\end{align}
where $\psi$ is an encoder function that applies the token mask $P_{m'}$ to textual embeddings as follows:
\setlength{\abovedisplayskip}{2pt}
\setlength{\belowdisplayskip}{4pt}
\begin{align}
\label{eq:psi}
\psi(T \mid P_{m'})
=
\frac{\bm{p}_{m'}}{\|\bm{p}_{m'}\|_{2}},
\quad
\bm{p}_{m'}
= \sum_{n=1}^{L} P_{m'n} \bm{t}_{n}.
\end{align}
These embeddings serve as phrase-level queries to identify corresponding image region subsets.
}

\subsection{Powerset Alignment}
\label{sec:powerset_alignment}

Powerset alignment establishes local-to-global alignment by minimizing a triplet margin loss defined based on the bidirectional similarity between region subsets $A$ and tree nodes $B$.
We first define the fine-grained similarity scores, and then aggregate them in two directions, R2T and T2R, as shown in Figure~\sref{fig:overview}{(c)}.

\paragraph{Fine-Grained Similarity.}
Let $\mathcal{B}=\{(I_{i}, T_{i})\}_{i=1}^{C}$ be a training mini-batch consisting of $C$ image-text pairs.
Given an image $I_{i}$ and a text description $T_{j}$ (potentially $j\!\not=\!i$),
we define their fine-grained similarity scores $Q_{i,j,A,B}$ by measuring inner products between the region embeddings $\bm{r}^{\smash{(i)}}_{A}$ and the phrase embeddings $\bm{p}_{B}^{(j)}$:
{
\setlength{\abovedisplayskip}{6pt}
\setlength{\belowdisplayskip}{6pt}
\begin{align}
\label{eq:Q}
Q_{i,j,A,B} = \langle \bm{r}^{(i)}_{A}, \bm{p}_{B}^{(j)} \rangle
\end{align}
where $A\!\subseteq\!\mathcal{M}_{i}$ is a region subset, $B\!\in\!\mathcal{T}_{j}$ is a tree node, and $i, j\!\in\!\{ 1, 2, \cdots, C\}$ index samples in the mini-batch.
}

\paragraph{R2T Aggregation.} This aggregation computes the best-matching phrase for each region subset and then aggregates corresponding scores by averaging.
Specifically, we define the R2T similarity matrix $Q^{\rightarrow}\hspace{-2pt}\!\in\!\mathbb{R}^{C \times C}$ as
{
\setlength{\abovedisplayskip}{5pt}
\setlength{\belowdisplayskip}{5pt}
\begin{align}
\label{eq:Q_right}
Q^{\rightarrow}_{i,j}
&=
\frac{1}{2^{M}}
\sum_{A \subseteq \mathcal{M}_{i}}
\max_{B \in \mathcal{T}_{j}}
Q_{i,j,A,B}.
\end{align}
This emphasizes region-level coverage, while neglecting less-relevant phrases that do not strongly correspond to any region subset.
}

\begin{figure}
\centering
\includegraphics[width=0.9\linewidth]{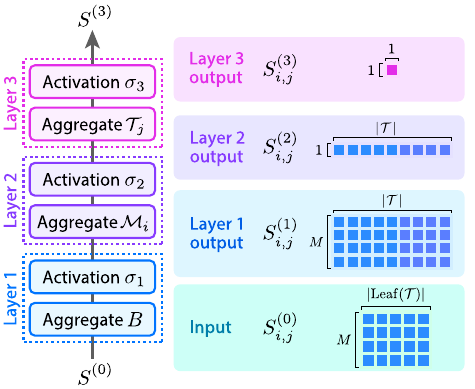}
\vspace{-10pt}
\caption{\textbf{Non-Linear Aggregator (NLA).} Each layer applies aggregation followed by activation.}
\label{fig:nla}
\vspace{-14pt}
\end{figure}

\paragraph{T2R Aggregation.}
Conversely, this aggregation computes the best-matching region subset for each phrase.
We define the T2R similarity matrix $Q^{\leftarrow}\!\in\!\mathbb{R}^{C \times C}$ as
{
\setlength{\abovedisplayskip}{6pt}
\setlength{\belowdisplayskip}{6pt}
\begin{align}
\label{eq:Q_left}
Q^{\leftarrow}_{i,j}
&=
\frac{1}{|\mathcal{T}_{j}|}
\sum_{B \in \mathcal{T}_{j}}
\max_{A \subseteq \mathcal{M}_{i}}
Q_{i,j,A,B}.
\end{align}
This emphasizes phrase-level grounding by ensuring each phrase is closely matched to a region subset.
}

\paragraph{Loss Function.}
Combining the two similarity matrices, we form the final similarity as $\bar{Q} = Q^{\rightarrow} + Q^{\leftarrow}$.
For training, we employ the triplet margin loss~\cite{Balntas2016Triplets, Asokan2025FineLIP} due to its effectiveness in encouraging margin-based discrimination between matched and mismatched pairs.
Specifically, our triplet loss is defined as
{
\setlength{\abovedisplayskip}{7pt}
\setlength{\belowdisplayskip}{7pt}
\begin{align}
\label{eq:triplet}
\mathcal{L}_{\text{triplet}}
=
\Phi_{\gamma}(\bar{Q}) + \Phi_{\gamma}(\bar{Q}^{\top})
\end{align}
where $\bar{Q}^{\top}$ is the transpose of $\bar{Q}$ and 
$\Phi_{\gamma}\!:\!\mathbb{R}^{C \times C}\!\to\!\mathbb{R}$ is the row-wise triplet loss function given by
\begin{align}
\Phi_{\gamma}(X)
= \frac{1}{C}
\sum_{i=1}^{C}
\max \left(
\max_{j \not= i} X_{i,j} - X_{i,i} + \gamma, 0
\right).
\end{align}
The final loss function is a sum of the CLIP contrastive loss and the triplet loss: $\mathcal{L}_{\text{total}}\!=\!\mathcal{L}_{\text{CLIP}}\!+\!\lambda \mathcal{L}_{\text{triplet}}$, where $\lambda\!=\!0.2$.
}

\paragraph{Discussion.}
Compared to token-to-token alignment frameworks \cite{Yao2022FILIP, Asokan2025FineLIP},
PowerCLIP establishes more exhaustive alignment.
However, computing the loss function poses a significant challenge, as it involves exponential complexity with respect to the number of region masks.
We address this limitation through theoretically grounded approximations.

\section{Tractable Approximations}

This section introduces \textbf{Non-Linear Aggregators (NLAs)}, which provide tractable approximations for the R2T and T2R aggregations. NLAs offer two primary advantages.
First, training stability is improved by employing soft assignment instead of hard assignment computed via max operations in Eqs.~(\ref{eq:Q_right}, \ref{eq:Q_left}).
Second, computational complexity of aggregation is significantly reduced from $\mathcal{O}(2^{M})$ to $\mathcal{O}(M)$ with respect to the number of masks $M$.

\subsection{General Form of NLAs}
The NLA comprises three layers, each consisting of an aggregation operation followed by an activation function, as shown in Figure~\ref{fig:nla}.
The input is the similarity tensor $S^{(0)}$,
obtained by computing inner products between individual region masks and phrases at leaf nodes:\footnote{Since the number of tree nodes depends on each textual description, $S^{(0)}$ is a pseudo tensor comprising $C M \times \sum_{j=1}^{C} |\mathrm{Leaf}(\mathcal{T}_{j})|$ scores.}
\begin{align}
S_{i, j, m, m'}^{(0)}
=
\langle
\phi(I_{i} \mid R_{m}), \psi(T_{j} \mid P_{m'})
\rangle,
\end{align}
where $R_{m}\!\in\!\mathcal{M}_{i}$ is a region mask for the image $I_{i}$ and $P_{m'}\!\in\!\mathrm{Leaf}(\mathcal{T}_{j})$ is a token mask at a leaf node for the description $T_{j}$.
The encoders $\phi, \psi$ are from Eqs.~(\ref{eq:phi}, \ref{eq:psi}).

At each layer, indexed by $l \in \{1,2,3\}$, the similarity scores in $S^{(l-1)}$ are aggregated by summation over a specific dimension followed by an optional activation function $\sigma_{l}\!:\!\mathbb{R}\!\to\!\mathbb{R}$.
Specifically, the first layer aggregates scores for each node $B \in \mathcal{T}_{j}$:
{
\setlength{\abovedisplayskip}{5pt}
\setlength{\belowdisplayskip}{5pt}
\begin{align}
S^{(1)}_{i,j,m | B}
=
\sigma_{1} \left(
\sum_{P_{m'} \in B} S_{i, j, m, m'}^{(0)} \right).
\end{align}
The second layer aggregates scores over region masks:
\begin{align}
S^{(2)}_{i,j | B}
=
\sigma_{2} \left(
\sum_{R_{m} \in \mathcal{M}_{i}} S^{(1)}_{i,j,m | B} \right).
\end{align}
Finally, the third layer aggregates scores over tree nodes:
\begin{align}
S_{i,j}^{(3)} =
\sigma_{3} \left(
\frac{1}{|\mathcal{T}_{j}|^{1-\alpha}}
\sum_{B \in \mathcal{T}_{j}} S^{(2)}_{i,j|B} \right),
\end{align}
where $\alpha \in [0,1]$ is a hyperparameter that interpolates between average aggregation ($\alpha=0$) and summation aggregation ($\alpha=1$).
}

These aggregation procedures avoid summation or maximization over powersets, thus reducing computational complexity. Nevertheless, the proposed design enables NLAs to approximate the R2T and T2R similarity matrices through a careful choice of activation functions.

\subsection{NLA-T1 for T2R Aggregation}

For the T2R aggregation, we introduce a specific class of NLAs, referred to as NLA Type 1 (NLA-T1).
Theorem 1 and Corollary 1 prove that NLA-T1 is a soft-assignment variant of the T2R aggregation, computing $Q^{\leftarrow}_{i,j}$ in Eq.~(\ref{eq:Q_left}). We provide a proof in Appendix A.

\vspace{2pt}
\paragraph{\textit{Definition 1 (NLA-T1).}}
\textit{NLA-T1 is a class of NLAs defined by the following activation functions and hyperparameters:}
\begin{align}
\label{eq:nla_t2r}
\sigma_{1}(x)
\!=\!
\tau \cdot \mathrm{Act}\!\left(\frac{x}{\tau}\right),\;
\sigma_{2} = \sigma_{3} = \mathrm{Id},\;
\alpha\!=\!0
\end{align}
\textit{where $\mathrm{Act}\!:\!\mathbb{R}\!\to\!\mathbb{R}$ is a nonlinear activation function, $\tau$ is a temperature hyperparameter, and $\mathrm{Id}$ is the identity function.}

\paragraph{\textit{Theorem 1.}} \textit{Suppose $\mathrm{Act}\!=\!\mathrm{Softplus}$. Then, NLA-T1 approximates the T2R similarity ${Q}^{\leftarrow}_{i,j}$ with arbitrary precision.
That is, for any $\epsilon > 0$, there exists $\tau > 0$ such that $| S_{i,j}^{(3)} - {Q}^{\leftarrow}_{i,j} | < \epsilon$.}

\vspace{4pt}
\paragraph{\textit{Corollary 1.}}
\textit{Suppose $\mathrm{Act}\!=\!\mathrm{ReLU}~(\text{i.e.,}~\tau \to 0)$. Then, NLA-T1 computes the exact T2R similarity ${Q}^{\leftarrow}_{i,j}$.}

\vspace{4pt}
\noindent 
As Corollary 1 is corresponding to the hard assignment in Eq.~(\ref{eq:Q_left}),
NLA-T1 using softplus with $\tau > 0$ can be interpreted as a soft assignment variant.
In practice, choosing small positive $\tau \simeq 0.001$ leads to improved performance.

\subsection{NLA-T2 for R2T Aggregation}

Approximating the R2T aggregation is relatively more challenging than approximating the T2R aggregation, as the summation operation is performed over the powerset.
Here, we introduce NLA Type 2 (NLA-T2), which evaluates the lower and upper bounds of the R2T similarity and interpolates between these bounds via a hyperparameter $\alpha \in [0,1]$.
Theorem 2 shows that NLA-T2 can approach the true similarity score arbitrarily closely by tuning $\alpha$. 

\vspace{2pt}
\paragraph{\textit{Definition 2 (NLA-T2).}}
\textit{NLA-T2 is a class of NLAs defined by the following activation functions and hyperparameters:}
\begin{align}
\hspace{-10pt}
\sigma_{1}(x)\!=\!\zeta_{\alpha}\!\left(\frac{x}{2\tau}\right),
\sigma_{2}(x) = \exp(x),\;
\sigma_{3}(x)\!=\!\tau \log(x),
\hspace{-10pt}
\end{align}
\textit{where $\zeta_{\alpha}(x) = x + \alpha \int \mathrm{Act}(x) \mathrm{d}x$ is a residual antiderivative of a differentiable activation function $\mathrm{Act}$, satisfying $\zeta_{\alpha}(0)=0$, and $\tau$ is a temperature hyperparameter.}

\vspace{2pt}
\paragraph{\textit{Theorem 2.}} \textit{Suppose $\mathrm{Act}=\tanh$. Then, NLA-T2 approximates the R2T similarity ${Q}^{\rightarrow}_{i,j}$ with arbitrary precision.
That is, for any $\epsilon>0$, there exist $\tau>0$ and $\alpha\in[0,1]$ such that $\big|S^{(3)}_{i,j}-{Q}^{\rightarrow}_{i,j}\big|<\epsilon$.}

\vspace{4pt}
\noindent We provide a proof in Appendix B. In practice, the lower bound ($\alpha \!=\!0$) and the upper bound ($\alpha\!=\!1$) are often close to each other when $\tau$ is small, making our approach robust to the choice of $\alpha$ (see Figure~\ref{fig:approximation} for analysis).

\subsection{Loss Function}
Finally, we approximate the triplet loss by replacing $\bar{Q}$ in Eq.~(\ref{eq:triplet}) with $\bar{S}$ obtained using the two types of NLAs:
\begin{align}
\bar{S} &= \operatorname{NLA-T1}(S^{(0)}) + \operatorname{NLA-T2}(S^{(0)}).
\end{align}
Compared with naive computations in Eqs~(\ref{eq:Q_left}, \ref{eq:Q_right}), this significantly reduces computational cost while maintaining or even improving performance.
\end{toggle}
\begin{toggle}
\begin{table*}[t]
\newlength{\numcolwidth}
\setlength{\numcolwidth}{9mm}
\newlength{\firstnumcolwidth}
\setlength{\firstnumcolwidth}{\numcolwidth}
\addtolength{\firstnumcolwidth}{1.5mm}
\newcolumntype{N}{>{\centering\arraybackslash}p{\numcolwidth}}
\newcolumntype{K}{>{\hspace{1.5mm}\centering\arraybackslash}p{\firstnumcolwidth}} %
\newcommand{\rothead}[1]{\makebox[0pt][l]{\hspace*{-3mm}\rotatebox[origin=l]{45}{\strut #1}}}
\centering
\small
\midsepremove
\setlength{\tabcolsep}{0pt}
\begin{tabular}{l*{1}{K@{\hspace{-0.6mm}}}*{16}{N@{\hspace{-0.6mm}}}N}
\toprule\\[-1.5em]
\raisebox{16pt}[0pt][0pt]{Method} &
\rothead{Food101~\cite{food101}} & \rothead{CIFAR10~\cite{cifar}} & \rothead{CIFAR100~\cite{cifar}} & \rothead{SUN397~\cite{sun397}} &
\rothead{Cars~\cite{cars}} & \rothead{VOC07~\cite{voc2007}} & \rothead{Aircraft~\cite{aircraft}} & \rothead{DTD~\cite{dtd}} &
\rothead{OxfordPets~\cite{pets}} & \rothead{Caltech101~\cite{caltech101}} & \rothead{Flowers~\cite{flowers}} & \rothead{STL10~\cite{stl10}} &
\rothead{EuroSAT~\cite{eurosat}} & \rothead{RESISC45~\cite{resisc45}} & \rothead{GTSRB~\cite{gtsrb}} & \rothead{Country~\cite{clip}} &
\rothead{PCam~\cite{pcam}} & \rothead{\textbf{Avg}} \\[-0.1em]
\midrule
CLIP~\cite{clip} & 42.3 & 57.7 & 25.0 & 44.1 & 17.0 & 50.5 & 1.7 & 16.5 & 53.9 & 73.5 & 26.0 & 82.0 & 18.7 & 26.5 & 9.4 & 4.5 & 48.0 & 35.1 \\ 
FLIP~\cite{Li2023flip} & 39.9 & 52.8 & 24.5 & 42.8 & 15.9 & 46.6 & 1.4 & 15.9 & 46.0 & 70.4 & 25.3 & 80.2 & 17.0 & 25.8 & 5.6 & 4.0 & 47.1 & 33.0 \\ 
A-CLIP~\cite{Yang2023a-clip} & 41.8 & 61.6 & 27.1 & 46.6 & 16.0 & 51.1 & 1.3 & 17.1 & 51.2 & 73.5 & 25.7 & 85.8 & 20.5 & 29.1 & 8.0 & 4.2 & 50.1 & 35.9 \\ 
E-CLIP~\cite{Wei2024e-clip} & 42.1 & 70.7 & 32.0 & 43.9 & 15.1 & 43.6 & 2.2 & 17.0 & 55.4 & \underline{73.7} & 28.4 & 85.6 & 22.9 & 30.0 & 9.6 & 4.7 & 50.0 & 36.9 \\ 
C-PGS~\cite{Pei2025CLIPPGS} & 46.5 & 73.5 & 37.3 & 47.5 & 19.9 & 55.1 & \textbf{3.1} & 19.8 & 58.1 & 72.7 & 30.7 & 88.2 & 22.8 & 30.4 & \textbf{10.9} & 4.5 & \underline{50.8} & 39.5 \\
FILIP~\cite{Yao2022FILIP} & 33.2 & 74.3 & 36.4 & 44.3 & 11.0 & 47.4 & 1.6 & 13.9 & 34.3 & 64.2 & 12.2 & \textbf{92.8} & \textbf{33.2} & 24.3 & 8.4 & 2.8 & 50.0 & 34.4 \\
SPARC~\cite{Bica2024SPARC} & 42.1 & 71.9 & 35.5 & 45.1 & 16.0 & \textbf{61.1} & 2.6 & 19.1 & 52.4 & 72.0 & 27.6 & 82.9 & 23.8 & 24.4 & \underline{9.8} & \underline{4.8} & 50.7 & 37.8 \\
\rowcolor{lightcyan} \textbf{PowerCLIP-R} & \underline{50.3} & \underline{74.7} & \textbf{43.5} & \underline{48.7} & \underline{22.9} & 53.2 & \underline{2.9} & \textbf{21.5} & \underline{58.7} & \textbf{75.7} & \underline{32.4} & 88.4 & \underline{30.8} & \textbf{37.5} & \underline{9.8} & 4.6 & 50.0 & \underline{41.5} \\ 
\rowcolor{lightcyan}\textbf{PowerCLIP-S} & \textbf{51.2} & \textbf{81.3} & \underline{40.1} & \textbf{50.5} & \textbf{23.5} & \underline{56.0} & 1.6 & \underline{21.3} & \textbf{61.0} & 72.9 & \textbf{32.5} & \underline{90.5} & 29.0 & \underline{33.9} & 7.8 & \textbf{5.4} & \textbf{59.7} & \textbf{42.2} \\
\bottomrule
\end{tabular}
\vspace{-0.2cm}
\caption{
\textbf{Zero-shot classification.} We report Top-1 accuracy (\%) for 17 diverse classification datasets. Avg indicates the average accuracy.
}
\label{tab:results_zero-shot_classification}
\end{table*}
\begin{table*}[t]
\centering
\small
\midsepremove
\setlength{\tabcolsep}{1.45pt}
\begin{tabular}{l|ccccccccc|ccccccccc|ccc}
\toprule
& \multicolumn{9}{c|}{Text Retrieval (Image to Text)} & \multicolumn{9}{c|}{Image Retrieval (Text to Image)} & \multicolumn{3}{c}{\multirow{2}{*}{Average}} \\[-0.1em]
Method & \multicolumn{3}{c}{MS-COCO} & \multicolumn{3}{c}{Flickr8K} & \multicolumn{3}{c|}{Flickr30K} & \multicolumn{3}{c}{MS-COCO} & \multicolumn{3}{c}{Flickr8K} & \multicolumn{3}{c|}{Flickr30K} & \\
\cmidrule(lr){2-4} \cmidrule(lr){5-7} \cmidrule(lr){8-10} \cmidrule(lr){11-13} \cmidrule(lr){14-16} \cmidrule(lr){17-19} \cmidrule(lr){20-22}
&
R\hspace{-1pt}@\hspace{-1pt}1 & R\hspace{-1pt}@\hspace{-1pt}5 & R\hspace{-1.2pt}@\hspace{-1.8pt}1\hspace{-0.2pt}0 &
R\hspace{-1pt}@\hspace{-1pt}1 & R\hspace{-1pt}@\hspace{-1pt}5 & R\hspace{-1.2pt}@\hspace{-1.8pt}1\hspace{-0.2pt}0 &
R\hspace{-1pt}@\hspace{-1pt}1 & R\hspace{-1pt}@\hspace{-1pt}5 & R\hspace{-1.2pt}@\hspace{-1.8pt}1\hspace{-0.2pt}0 &
R\hspace{-1pt}@\hspace{-1pt}1 & R\hspace{-1pt}@\hspace{-1pt}5 & R\hspace{-1.2pt}@\hspace{-1.8pt}1\hspace{-0.2pt}0 &
R\hspace{-1pt}@\hspace{-1pt}1 & R\hspace{-1pt}@\hspace{-1pt}5 & R\hspace{-1.2pt}@\hspace{-1.8pt}1\hspace{-0.2pt}0 &
R\hspace{-1pt}@\hspace{-1pt}1 & R\hspace{-1pt}@\hspace{-1pt}5 & R\hspace{-1.2pt}@\hspace{-1.8pt}1\hspace{-0.2pt}0 &
R\hspace{-1pt}@\hspace{-1pt}1 & R\hspace{-1pt}@\hspace{-1pt}5 & R\hspace{-1.2pt}@\hspace{-1.8pt}1\hspace{-0.2pt}0 \\
\midrule
CLIP~\cite{clip} & 34.6 & 62.0 & 72.7 & 55.7 & 81.6 & 89.9 & 58.5 & 83.8 & 89.1 & 23.5 & 47.8 & 59.7 & 40.5 & 68.9 & 80.2 & 43.2 & 70.4 & 80.4 & 42.7 & 69.1 & 78.7 \\
FLIP~\cite{Li2023flip} & 32.6 & 59.1 & 70.6 & 55.0 & 80.9 & 88.9 & 53.8 & 80.8 & 88.5 & 22.6 & 46.1 & 58.1 & 40.3 & 68.1 & 78.6 & 41.5 & 67.9 & 77.5 & 41.0 & 67.1 & 77.0 \\
A-CLIP~\cite{Yang2023a-clip} & 33.7 & 60.2 & 71.0 & 53.7 & 80.1 & 88.0 & 55.3 & 81.4 & 87.6 & 23.9 & 48.3 & 60.0 & 40.6 & 68.9 & 78.9 & 43.1 & 70.1 & 78.8 & 41.7 & 68.2 & 77.4 \\
E-CLIP~\cite{Wei2024e-clip} & 34.3 & 62.0 & 73.3 & 57.0 & 82.7 & 90.1 & 55.8 & 84.2 & 89.6 & 23.8 & 48.2 & 59.8 & 42.0 & 69.4 & 79.6 & 43.3 & 70.9 & 80.2 & 42.7 & 69.6 & 78.8 \\
C-PGS~\cite{Pei2025CLIPPGS} & 36.0 & \underline{64.4} & 74.6 & 58.3 & 82.9 & 90.8 & 59.9 & 83.5 & 90.8 & 25.1 & 49.5 & 61.6 & 44.4 & 71.7 & 81.1 & \underline{47.1} & 73.5 & 82.0 & 45.1 & 70.9 & 80.1 \\
FILIP~\cite{Yao2022FILIP} & 16.8 & 38.0 & 50.8 & 31.2 & 55.2 & 66.8 & 35.7 & 61.0 & 72.5 & 14.0 & 33.3 & 44.8 & 24.2 & 50.2 & 62.3 & 27.3 & 55.1 & 65.8 & 24.9 & 48.8 & 60.5 \\
SPARC~\cite{Bica2024SPARC} & 33.7 & 60.9 & 72.3 & 55.2 & 82.2 & 90.5 & 57.1 & 82.6 & 89.6 & 23.8 & 48.0 & 59.6 & 41.0 & 70.1 & 79.3 & 42.7 & 71.3 & 80.1 & 42.3 & 69.2 & 78.6 \\
\rowcolor{lightcyan} \textbf{PowerCLIP-R} & \underline{36.7} & 64.0 & \underline{75.0} & \underline{58.5} & \textbf{84.8} & \underline{91.4} & \underline{61.7} & \underline{84.8} & \underline{91.9} & \underline{26.3} & \underline{51.1} & \underline{62.7} & \underline{44.8} & \underline{72.7} & \underline{82.4} & 46.6 & \underline{74.3} & \underline{82.7} & \underline{45.8} & \underline{72.0} & \underline{81.0} \\
\rowcolor{lightcyan} \textbf{PowerCLIP-S} & \textbf{37.3} & \textbf{64.9} & \textbf{75.6} & \textbf{58.6} & \underline{84.4} & \textbf{91.5} & \textbf{62.4} & \textbf{88.5} & \textbf{94.2} & \textbf{27.0} & \textbf{52.9} & \textbf{64.0} & \textbf{46.3} & \textbf{74.1} & \textbf{83.2} & \textbf{50.4} & \textbf{76.6} & \textbf{84.6} & \textbf{47.0} & \textbf{73.6} & \textbf{82.2} \\
\bottomrule
\end{tabular}
\vspace{-0.2cm}
\caption{\textbf{Zero-shot image-text retrieval}.
R\hspace{-1pt}@\hspace{-1pt}$K$ indicates recall (\%) at top $K = 1, 5,$ and $10$. Average columns are means across the six settings (MS-COCO, Flickr8K, Flickr30K for both Text Retrieval and Image Retrieval).}
\label{tab:results_zero-shot_retrieval}
\vspace{-0.3cm}
\end{table*}

\begin{table*}[ht]
\centering
\begin{minipage}{0.73\linewidth}
\centering
\midsepremove
\small
\setlength{\tabcolsep}{2pt}
\begin{tabular}{l|cccccc|ccc}
\toprule
Method & \scalebox{0.97}[1.0]{ImgNet-1k} & \scalebox{0.97}[1.0]{ImgNet-V2} & \scalebox{0.97}[1.0]{ImgNet-A} & \scalebox{0.97}[1.0]{ImgNet-R} & \scalebox{0.97}[1.0]{ImgNet-O} & \scalebox{0.97}[1.0]{ImgNet-S} & IN & OOD & All \\
\midrule
CLIP~\cite{clip}     & 36.1 & 30.7 & 8.0 & 47.6 & 38.4 & 24.9 & 36.1 & 29.0 & 31.0 \\ 
FLIP~\cite{Li2023flip}     & 34.4 & 29.5 & 7.1 & 41.4 & 39.5 & 20.1 & 34.4 & 27.5 & 28.7 \\ 
A-CLIP~\cite{Yang2023a-clip} & 35.2 & 30.1 & 8.1 & 45.1 & 39.4 & 23.7 & 35.2 & 30.3 & 30.3 \\ 
E-CLIP~\cite{Wei2024e-clip} & 36.3 & 30.7 & 8.1 & 47.9 & 39.6 & 25.4 & 36.3 & 30.3 & 31.3 \\
C-PGS~\cite{Pei2025CLIPPGS} & 38.6 & 33.1 & 9.6 & 48.1 & \textbf{42.6} & 25.6 & 38.6 & 31.8 & 32.9 \\
FILIP~\cite{Yao2022FILIP} & 26.7 & 22.9 & 9.3 & 37.4 & 25.9 & 18.2 & 26.7 & 22.7 & 23.4 \\
SPARC~\cite{Bica2024SPARC} & 37.2 & 32.1 & 9.3 & 46.8 & \underline{42.2} & 24.5 & 37.2 & 31.0 & 32.0 \\
\rowcolor{lightcyan} \scalebox{0.97}[1.0]{PowerCLIP-R} & \underline{40.3} & \underline{34.8} & \underline{11.2} & \underline{53.2} & 40.2 & \underline{28.7} & \underline{40.3} & \underline{33.6} & \underline{34.7} \\
\rowcolor{lightcyan} \scalebox{0.97}[1.0]{PowerCLIP-S} & \textbf{40.8} & \textbf{35.1} & \textbf{11.9} & \textbf{53.5} & 40.5 & \textbf{28.9} & \textbf{40.8} & \textbf{34.0} & \textbf{35.1} \\
\bottomrule
\end{tabular}
\vspace{-0.2cm}
\caption{
\textbf{Robustness evaluation.}
Top-1 accuracy for six ImageNet (ImgNet) datasets are reported with in-distribution (ID), out-of-distribution (OOD) and overall (All) averages.
}
\label{tab:results_robustness}
\end{minipage}
\hfill
\begin{minipage}{0.25\linewidth}
\centering
\midsepremove
\small
\setlength{\tabcolsep}{3pt}
\begin{tabular}{l|ccc}
\toprule
Method & Obj & Att & Rel \\
\midrule
CLIP~\cite{clip} & 73.9 & 68.8 & 64.5 \\
FLIP~\cite{Li2023flip} & 72.0 & 66.9 & 66.0 \\
A-CLIP~\cite{Yang2023a-clip} & 70.2 & 68.5 & 63.2 \\
E-CLIP~\cite{Wei2024e-clip} & 73.2 & 67.9 & 60.2 \\
C-PGS~\cite{Pei2025CLIPPGS} & 75.5 & \textbf{70.8} & \textbf{67.9} \\
FILIP~\cite{Yao2022FILIP} & 64.9 & 58.2 & 56.8 \\
SPARC~\cite{Bica2024SPARC} & 73.5 & \underline{70.4} & 66.9 \\
\rowcolor{lightcyan} \scalebox{0.97}[1.0]{PowerCLIP-R} & 75.6 & 70.3 & \textbf{67.9} \\
\rowcolor{lightcyan} \scalebox{0.97}[1.0]{PowerCLIP-S} & \textbf{76.1} & \underline{70.4} & 67.1 \\
\bottomrule
\end{tabular}
\vspace{-0.2cm}
\caption{
\textbf{Compositionality evaluation on SugarCrepe.}
}
\label{tab:results_compositionality}
\end{minipage}
\end{table*}

\def\clsours{42.2}
\def\retours{47.0}
\begin{table*}[t]
\centering
\small
\begin{minipage}{0.30\textwidth}
\vspace{-6pt}
\centering
\setlength{\tabcolsep}{5pt}
{
\midsepremove
\renewcommand{\arraystretch}{0.957}
\begin{tabular}{l|ccc}
\toprule
Method
& Text & Image & Group \\
\midrule
CLIP~\cite{clip} & \underline{24.8} & 8.0 & 4.3 \\
FLIP~\cite{Li2023flip} & \underline{24.8} & 10.0 & 5.8 \\
C-PGS~\cite{Pei2025CLIPPGS} & \textbf{25.2} & 10.5 & 7.2 \\
FILIP~\cite{Yao2022FILIP} & 21.3 & \underline{13.5} & \underline{9.7} \\
SPARC~\cite{Bica2024SPARC} & 23.3 & 12.7 & 9.0\\
\rowcolor{lightcyan} PowerCLIP-R & 22.5 & 9.5 & 6.5 \\
\rowcolor{lightcyan} PowerCLIP-S & \underline{24.8} & \textbf{16.0} & \textbf{10.2} \\
\bottomrule
\end{tabular}
}
\vspace{-0.2cm}
\caption{
\textbf{Compositionality evaluation on Winoground.}
}
\label{tab:results_winoground}
\vspace{-0.2cm}
\setlength{\tabcolsep}{2pt}
\end{minipage}
\hfill
\begin{minipage}{0.20\textwidth}
\centering
{
\renewcommand{\arraystretch}{1.065}
\setlength{\tabcolsep}{2pt}
\midsepremove
\begin{tabular}{lcc}
\toprule
Method & Cls & Ret\\
\midrule
PowerCLIP-S & \clsours & \retours \\
w/o region sets & 41.1 & 45.7\\
w/o parse trees & 41.1 & 45.4 \\
w/o R2T agg. & 40.8 & 45.3 \\
w/o T2R agg. & 41.8 & 45.4 \\
w/o Triplet loss & 35.1 & 42.7 \\
\bottomrule
\end{tabular}
\vspace{-0.2cm}
\caption{Ablation study for key components.}
\label{tab:ablation}
}
\end{minipage}
\hfill
\begin{minipage}{0.20\textwidth}
\centering
\setlength{\tabcolsep}{3pt}
\vspace{-1pt}
{
\renewcommand{\arraystretch}{1.06}
\midsepremove
\begin{tabular}{lccc}
\toprule
Mask & $M$ & Cls & Ret\\
\midrule
Random & 5 & 40.5 & 45.5 \\
Random & 10 & 41.5 & 45.8 \\
Random & 15 & 40.9 & 43.9 \\
\midrule
SAM & 5 & 41.3 & 45.2 \\
SAM & 10 & \clsours & \retours \\
SAM & 15 & 41.4 & 44.9 \\
\bottomrule
\end{tabular}
\vspace{-0.2cm}
\caption{Mask generation. $M$: Number of masks.}
\label{tab:mask}
}
\end{minipage}
\hfill
\begin{minipage}{0.23\textwidth}
\centering
\setlength{\tabcolsep}{2pt}
\midsepremove
\begin{tabular}{ccccc}
\toprule
NLA-T1 & NLA-T2 & Cls & Ret \\
\midrule
Softplus & Tanh & \clsours & \retours \\
ReLU & Tanh & 40.2 & 45.0\\
GELU & Tanh & 41.8 & 44.9\\
Swish & Tanh & 41.1 & 45.3 \\
\midrule
Softplus & Tanh & \clsours & \retours \\
Softplus & Sigmoid & 40.5 & 45.0 \\
Softplus & SoftSign & 41.0 & 44.9 \\
\bottomrule
\end{tabular}
\vspace{-1pt}
\caption{Activation functions.}
\label{tab:activation}
\end{minipage}
\end{table*}

\section{Experiments}

\subsection{Experimental Setting}

\paragraph{Datasets and Tasks.}
Following \cite{Wei2024e-clip, Pei2025CLIPPGS}, we use the Conceptual Captions 12M (CC12M) dataset~\cite{changpinyo2021cc12m} for training.
For extensive evaluation, models are evaluated across 28 benchmarks.
Specifically, we conduct evaluation on
(i) 17 diverse datasets for zero-shot classification (listed in Table~\ref{tab:results_zero-shot_classification}),
(ii) 3 datasets for image-text retrieval (COCO~\cite{Lin2014COCO}, Flickr8k~\cite{Young2014Flickr} and Flickr30k~\cite{Young2014Flickr}),
(iii) 6 datasets for robustness evaluation (ImageNet-1k~\cite{Deng2009ImageNet}, -V2~\cite{Recht2019ImageNetV2}, -A~\cite{Hendrycks2021ImageNetAO}, -R~\cite{Hendrycks2021ImageNetR}, -O~\cite{Hendrycks2021ImageNetAO} and -Sketch~\cite{Wang2019ImageNetSketch}),
and 
(iv) 2 datasets for compositionality evaluation (SugarCrepe~\cite{Hsieh2023SugarCrepe} and Winoground~\cite{Thrush2022Winoground}).

\paragraph{Baselines.}
We compare PowerCLIP with seven baselines: CLIP~\cite{clip}, FLIP~\cite{Li2023flip}, A-CLIP~\cite{Yang2023a-clip}, E-CLIP~\cite{Wei2024e-clip},
C-PGS~\cite{Pei2025CLIPPGS}, FILIP~\cite{Yao2022FILIP}, and SPARC~\cite{Bica2024SPARC}.
These baselines cover different global and local alignment strategies. All models are evaluated under a consistent setting.

\paragraph{Implementation.}
We adopt the training setting of~\cite{Wei2024e-clip, Pei2025CLIPPGS}.
Specifically, ViT-B/16~\cite{Dosovitskiy2020ViT} is used as the image encoder, with images resized to a $224\!\times\!224$. The text encoder is a Transformer consisting of 12 layers, 8 attention heads, and embedding dimensions of 512. The models are trained for 32 epochs using the AdamW optimizer with a cosine decay learning rate scheduler, an initial learning rate of $10^{-3}$, weight decay of 0.2, and a batch size of 4,096.
The number of masks $M$ is set to 10.
We use softplus and tanh activations with $\tau=0.001$ for NLA-T1 and NLA-T2, respectively.
For NLA-T2, the function $\zeta_{\alpha}(x)$ is given by $\log \cosh (x)$ and $\alpha$ is set to 0.75.
We implement two variants of our approach: PowerCLIP-R, which uses random masks, and PowerCLIP-S, which uses masks randomly selected from those generated by SAM2~\cite{ravi2024sam2}.

\subsection{Experimental Results}

\paragraph{Zero-Shot Classification.}
Table~\ref{tab:results_zero-shot_classification} summarizes zero-shot classification results.
We observe that PowerCLIP-R significantly outperforms the CLIP baseline (+6.4\%), while PowerCLIP-S further improves the performance, achieving the best average accuracy of 42.2\% across 17 datasets.
Significant gains are observed on challenging fine-grained datasets such as Cars (+6.5\%), Food101 (+8.9\%), and RESISC45 (+7.4\%).
Compared to state-of-the-art methods for global alignment (C-PGS) and local alignment (SPARC), PowerCLIP-S achieves +2.7 and +4.4 points higher average accuracy, respectively, surpassing both on 14 out of 17 datasets.
These results demonstrate the superiority of our local-to-global alignment in capturing nuanced semantics.

\paragraph{Zero-Shot Image-Text Retrieval.}
Table~\ref{tab:results_zero-shot_retrieval} presents zero-shot retrieval results.
PowerCLIP achieves consistent improvements over baseline methods, surpassing CLIP with an average gain of +4.3\% for Recall@1 across both retrieval tasks.
Notably, PowerCLIP surpasses baselines across all retrieval scenarios.
These results demonstrate the effectiveness of compositional alignment between textual phrases and image region combinations in retrieval tasks.

\paragraph{Robustness.}
Table~\ref{tab:results_robustness} compares PowerCLIP with baselines across six ImageNet robustness benchmarks.
PowerCLIP significantly surpasses the baselines in terms of both in-distribution (ID) and out-of-distribution (OOD) average accuracy. Particularly notable is its performance on ImageNet-R (+5.9\%) and ImageNet-Sketch (+4.0\%), datasets designed to assess robustness under domain shifts.
Overall, the results underscore the generalizability and robustness of PowerCLIP in challenging scenarios.

\paragraph{Compositionality.}
Tables~\ref{tab:results_compositionality} and \ref{tab:results_winoground} evaluate compositional understanding on SugarCrepe (average accuracies for object, attribute and relation subsets) and Winoground (text, image and overall group accuracy), respectively.
Consistent with other evaluations, PowerCLIP significantly improves average accuracy over CLIP, confirming stronger compositional grounding of novel elements introduced in images. 
Performance improvements are particularly pronounced for the object subset of SugarCrepe (+2.2\%) and for image retrieval on Winoground (+8.0\%).
These results demonstrate that explicit phrase-to-region alignment enhances fine-grained compositional understanding, aligning precisely with our motivation.

\begin{figure*}
\centering
\includegraphics[width=\linewidth]{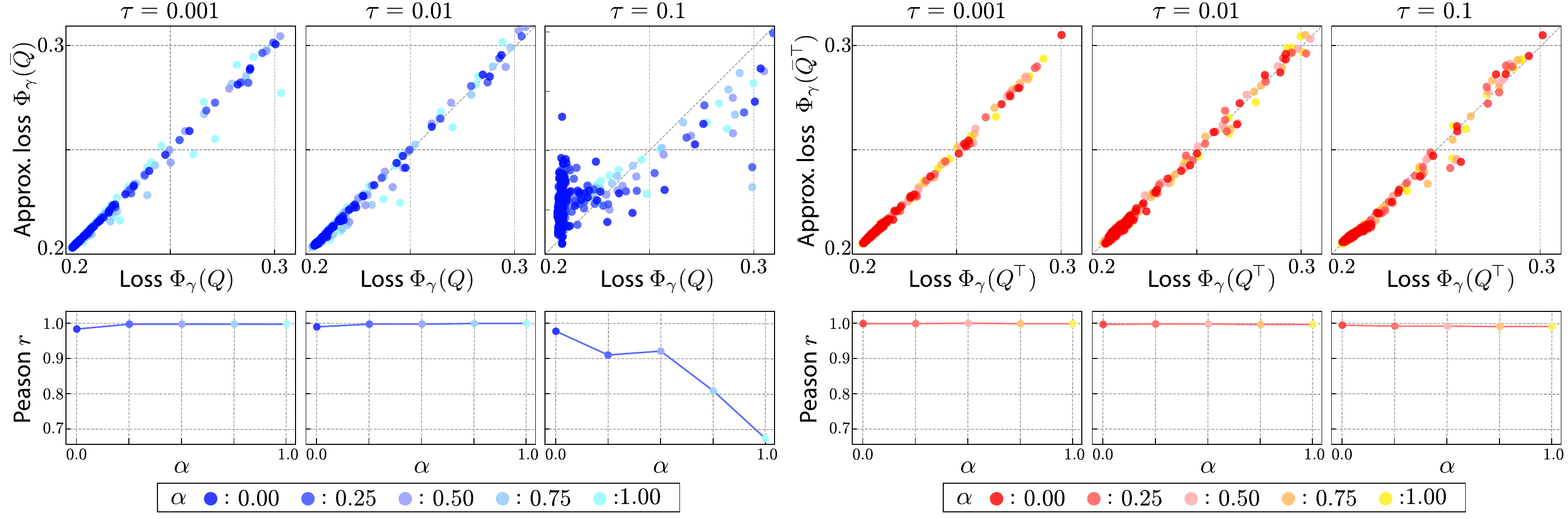}
\vspace{-14pt}
\caption{
Approximation accuracy evaluation.
Top: Comparison between exact and approximated losses for $\tau = \{0.1, 0.01, 0.001\}$ and $\alpha \in \{0.00, 0.25, 0.50, 0.75, 1.00\}$.
Bottom: Pearson correlation $r$ between exact and approximated losses.
}
\label{fig:approximation}
\vspace{-3pt}
\end{figure*}

\begin{figure*}
\centering
\includegraphics[width=\linewidth,height=3cm]{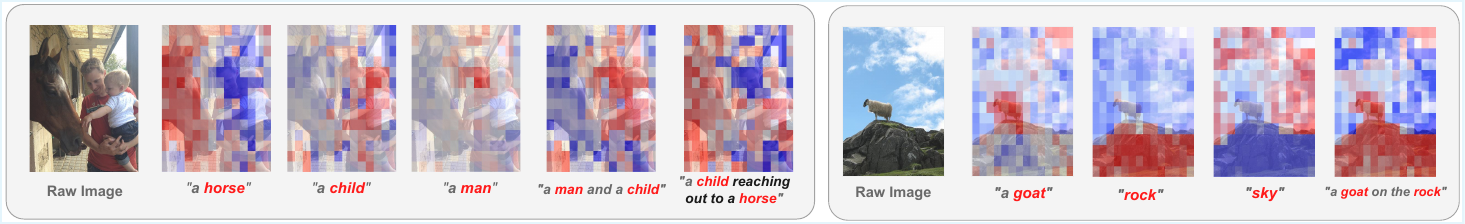}
\captionof{figure}{Visualizations of text-to-patch similarities.
For each input text, we compute similarities between the text representation and image patch features and visualize them as heatmaps, showing that high responses are concentrated on the regions referred to in the text.}
\label{fig:qualitative}
\vspace{-3pt}
\end{figure*}

\subsection{Analysis and Discussion}
\paragraph{Ablation Study.}
Table~\ref{tab:ablation} quantifies the contributions of key PowerCLIP components through systematic ablations: 1) replacing region sets with individual regions, 2) replacing parse trees with individual tokens, 3) omitting the R2T aggregation loss, 4) omitting the T2R aggregation loss, and 5) omitting the proposed triplet loss. Results confirm that each component contributes to the overall performance, underscoring their complementary roles.

\paragraph{Mask Generation.} Table~\ref{tab:mask} investigates mask-generation methods by varying the number ($M$) and type of masks. 
SAM-generated masks achieve higher performance than random masks overall, with the best results obtained at $M=10$.
Random masks also maintain performance in the same range and do not break down when a sufficient number of masks is used.
These results suggest that our method is relatively robust to both the mask generation strategy and the number of masks, while still benefiting from a modest performance gain when using SAM.

\paragraph{Activation Functions.} 
Table~\ref{tab:activation} compares activation functions for NLAs.
For NLA-T1, Softplus consistently outperforms ReLU, GELU and Swish due to its smooth approximation of max operations essential for T2R aggregation.
For NLA T2, Tanh performs the best for approximating R2T.
Sigmoid and SoftSign still capture the mapping but show clear drops in performance compared with Tanh, similar in scale to the differences seen in NLA T1.
Thus, Tanh is the most suitable choice for NLA T2, while other smooth activations remain usable but less accurate.

\begin{table}[t]
\begin{minipage}[t]{0.48\linewidth}
\centering
\small
\midsepremove
\setlength{\tabcolsep}{2pt}
\begin{tabular}{lccc}
\toprule
Method & CF10 & CF100 & IN1k \\[-0.1em]
\midrule
CLIP~\cite{clip}     & 88.0 & 67.4 & 62.3  \\ 
FLIP~\cite{Li2023flip}     & 85.9 & 65.5 & 61.3  \\ 
A-CLIP~\cite{Yang2023a-clip} & 86.4 & 66.1 & 62.0 \\ 
E-CLIP~\cite{Wei2024e-clip} & 89.0 & \underline{69.7} & 62.7 \\ 
FILIP~\cite{Yao2022FILIP} & 84.4 & 56.8 & 50.4 \\
SPARC~\cite{Bica2024SPARC} & 88.7 & 69.4 & 62.7\\
C-PGS~\cite{Pei2025CLIPPGS} & \underline{90.0} & \textbf{72.3} & \underline{64.4} \\
\rowcolor{lightcyan} PowerCLIP & \textbf{91.3} & \textbf{72.3} & \textbf{65.8} \\
\bottomrule
\end{tabular}
\caption{Linear probing.}
\label{tab:linear_probing}
\end{minipage}
\hfill
\begin{minipage}[t]{0.48\linewidth}
\centering
\small
\setlength{\tabcolsep}{4pt}
{
\midsepremove
\begin{tabular}{lcc}
\toprule
Value & Cls & Ret\\
\midrule
$\alpha\!=\!0.00$ & 40.7 & 49.0\\
$\alpha\!=\!0.25$ & 41.8 & 47.9\\
$\alpha\!=\!0.50$ & 42.3 & 49.0\\
$\alpha\!=\!0.75$ & 42.2 & 47.0\\
$\alpha\!=\!1.00$ & 42.0 & 47.4\\
\midrule
$\tau\!=\!0.001$ & 42.2 & 47.0\\
$\tau\!=\!0.01$ & 41.7 & 46.5\\
$\tau\!=\!0.1$ & 40.7 & 45.6\\
\bottomrule
\end{tabular}
\caption{Hyperparameter.}
\label{tab:hyperparameter}
}
\end{minipage}
\vspace{-14pt}
\end{table}

\paragraph{Linear Probing.} 
Table~\ref{tab:linear_probing} presents linear probing results on CIFAR10 (CF10), CIFAR100 (CF100), and ImageNet-1k (IN1k). Consistent with zero-shot evaluations, PowerCLIP achieves the best performance. This indicates that PowerCLIP learns more discriminative features, enabling improved linear separability for classification tasks.

\paragraph{NLA Accuracy.} 
Figure~\ref{fig:approximation} analyzes the approximation accuracy of NLAs for the two triplet loss terms $\Phi_{\gamma}(Q)$ and $\Phi_{\gamma}(Q^{\top})$.
We observe that loss values approximated by NLAs closely match the exact values when $\tau$ is small (0.001 or 0.01), consistently achieving Pearson correlations above 0.98 across all tested $\alpha$.
The highest correlation (0.999) was obtained with $\tau=0.001$ and $\alpha=0.75$.
Table~\ref{tab:hyperparameter} shows that performance stays high for $\alpha>0.5$, and peaks at $\alpha=0.75$.
Similarly, $\tau$ follows the same tendency predicted by our theoretical analysis, with smaller values leading to better performance.

\paragraph{Qualitative Examples}
Figure~\ref{fig:qualitative} provides qualitative examples.
For the illustrated examples, PowerCLIP produces text-to-patch similarity heatmaps whose high responses are concentrated on image regions corresponding to words explicitly mentioned in the text. Across different prompts, the highlighted patches consistently align with the referred objects and actions, indicating that the model attends to the intended visual evidence rather than unrelated areas. 

We include computational cost comparisons in Sec.~\ref{sec:appendix_cost}.

\end{toggle}

\section{Conclusion}
We introduced PowerCLIP,
a novel contrastive pre-training framework that leverages powerset alignment.
PowerCLIP exhaustively optimizes local-to-global alignments by minimizing bidirectional triplet losses defined over the powersets of image regions and textual parse trees.
Extensive experimental results demonstrate that PowerCLIP achieves state-of-the-art performance across diverse benchmarks.
For future work, extending PowerCLIP to 3D scene understanding presents a promising avenue for enhancing spatial and semantic alignment in more complex multimodal scenarios.

\section{Acknowledgements}
This work was supported by Japan Science and Technology Agency (JST) as part of Adopting Sustainable Partnerships for Innovative Research Ecosystem (ASPIRE), Grant Number JPMJAP2518. This work was supported by the AIST policy-based budget project ``R\&D on Generative AI Foundation Models for the Physical Domain''. We used ABCI 3.0 provided by AIST and AIST Solutions with support from ``ABCI 3.0 Development Acceleration Use''. We would like to thank Yukito Tajima and Daisuke Nohara for their valuable support with the implementation of this work.

{
    \small
    \bibliographystyle{ieeenat_fullname}
    \bibliography{main}
}

\begin{toggle}

\clearpage
\setcounter{page}{1}
\maketitlesupplementary
\appendix
\section*{Appendix A. Proof of Theorem 1}

In this section, we present a proof of Theorem 1. We first restate the definitions of the T2R aggregation and NLAs.

\subsection*{A.1 Preliminary}
\paragraph{Notation.}
Let $\mathcal{M}_{i}\!=\!\{R_m\}_{m=1}^{M}$ be the set of $M$ region masks for the image $I_{i}$,
and let $\mathcal{T}_{j}$ be the parse tree for the text description $T_{j}$ with $K_{j}\!=\!|\mathcal{T}_{j}|$ nodes, where $i$ and $j$ index samples in mini-batches. For token masks $P_{m'} \in \mathrm{Leaf}(\mathcal{T}_{j})$, we define
\begin{align}
S^{(0)}_{i,j,m,m'} := \big\langle \phi(I_i\!\mid R_m),\,\psi(T_j\!\mid P_{m'}) \big\rangle,
\end{align}
where $\phi$ and $\psi$ are image and text encoders, respectively, satisfying $\|\phi(\cdot)\|_2=\|\psi(\cdot)\|_2=1$ so that $|S^{(0)}_{i,j,m,m'}| \le 1$.
For any node $B\in\mathcal{T}_j$, we define aggregated similarities:
\begin{align}
Q_{i,j,m,B} &:= \sum_{P_{m'}\in B} S^{(0)}_{i,j,m,m'},\\
\label{appeq:Q_ijAB}
Q_{i,j,A,B} &:= \sum_{R_m\in A} Q_{i,j,m,B},\\\bar{Q}_{i,j,B} &:= \sum_{R_m\in \mathcal{M}_i} Q_{i,j,m,B}.
\end{align}
Note that, we have $Q_{i,j,A,B}\!=\!\langle \bm{r}^{(i)}_A,\,\bm{p}^{(j)}_B \rangle$ by bilinearity.

\paragraph{T2R Aggregation.}
Given similarity $Q_{i,j,A,B}$, we define the T2R similarity as
\begin{align}
Q^{\leftarrow}_{i,j} := \frac{1}{K_{j}} \sum_{B\in\mathcal{T}_j} Q^{\leftarrow}_{i,j,B},
\end{align}
where $Q^{\leftarrow}_{i,j,B}$ evaluates the best-matching region subset $A$ for each node $B$ as
\begin{align}
\label{appeq:O_ijB}
Q^{\leftarrow}_{i,j,B} = \max_{A\subseteq\mathcal{M}_i} Q_{i,j,A,B}.
\end{align}

\paragraph{Non-Linear Aggregators (NLAs).}
Given $S^{(0)}$, we define the three-layer NLAs with a hyperparameter $\alpha\in[0,1]$ and activation functions $\{\sigma_{l}\}_{l=1}^{3}$:
\vspace{-5pt}
\begin{align}
S^{(1)}_{i,j,m| B} &:= \sigma_{1}\left(\sum_{P_{m'}\in B} S^{(0)}_{i,j,m,m'}\right),\\
S^{(2)}_{i,j| B} &:= \sigma_{2}\!\left(\sum_{R_m\in\mathcal{M}_i} S^{(1)}_{i,j,m| B}\right),\\
S^{(3)}_{i,j} &:= \sigma_{3}\!\left(\frac{1}{K_{j}^{\,1-\alpha}} \sum_{B\in\mathcal{T}_j} S^{(2)}_{i,j| B}\right).
\end{align}

\paragraph{\textit{Lemma A.1 (LSE Bound).}}
\textit{Given aggregated similarities $Q_{i,j,A,B}$ and $Q^{\leftarrow}_{i,j,B}$, for any $\tau>0$, we have}
\begin{align}
\left|\;
\tau \log\hspace{-5pt}\sum_{A \subseteq \mathcal{M}_{i}}\hspace{-5pt}\exp\!\left(\frac{Q_{i,j,A,B}}{\tau}\right)
- Q^{\leftarrow}_{i,j,B}\;\right|
\leq \tau M \log 2.
\end{align}
\begin{proof}
Considering the log-sum-exp (LSE) bound, the Lemma immediately holds; that is, we have 
\begin{align}
\hspace{-5pt}
&
\left| 
\tau \log\hspace{-5pt}\sum_{A \subseteq \mathcal{M}_{i}}\hspace{-5pt}\exp\!\left(\frac{Q_{i,j,A,B}}{\tau}\right)
- Q^{\leftarrow}_{i,j,B} \right| \\
&\leq
\tau \log \sum_{A \subseteq \mathcal{M}_{i}} \exp\!\left(\frac{Q_{i,j,A,B}-Q^{\leftarrow}_{i,j,B}}{\tau}\right) \\
&\leq
\tau \log \sum_{A \subseteq \mathcal{M}_{i}} 1 \\
&= \tau \log \left|2^{\mathcal{M}_{i}}\right|\\
&= \tau M \log 2.
\end{align}
This completes the proof.
\end{proof}

\subsection*{A.2 NLA-T1}

We define NLA-T1 and provide a proof for Theorem 1.

\paragraph{\textit{Definition 1 (NLA-T1).}}
\textit{NLA-T1 is a class of NLAs defined by the following activation functions and hyperparameters:}
\begin{align}
\label{app_eq:nla_t2r}
\sigma_{1}(x)
\!=\!
\tau \cdot \mathrm{Act}\!\left(\frac{x}{\tau}\right),\;
\sigma_{2} = \sigma_{3} = \mathrm{Id},\;
\alpha\!=\!0
\end{align}
\textit{where $\mathrm{Act}\!:\!\mathbb{R}\!\to\!\mathbb{R}$ is a nonlinear activation function, $\tau$ is a temperature hyperparameter, and $\mathrm{Id}$ is the identity function.}

\paragraph{\textit{Theorem 1.}} \textit{Suppose $\mathrm{Act}\!=\!\mathrm{Softplus}$. Then, NLA-T1 approximates the T2R similarity ${Q}^{\leftarrow}_{i,j}$ with arbitrary precision.
That is, for any $\epsilon > 0$, there exists $\tau > 0$ such that $| S_{i,j}^{(3)} - {Q}^{\leftarrow}_{i,j} | < \epsilon$.}
\begin{proof}
From Definition 1, the output for the second layer of NLA-T1 is given by
\begin{align}
S^{(2)}_{i,j | B}
=
\sum_{R_{m} \in \mathcal{M}_{i}} \tau \cdot \mathrm{Act} \left( \sum_{P_{m'} \in B} \frac{S_{i, j, m, m'}^{(0)}}{\tau}\right).
\end{align}
When the softplus function is used for the activation function, we have
\begin{align}
\hspace{-5pt}
S^{(2)}_{i,j | B}
&=\hspace{-2pt}\sum_{R_{m} \in \mathcal{M}_{i}} \tau \cdot \mathrm{Softplus} \left( \sum_{P_{m'} \in B} \frac{S_{i, j, m, m'}^{(0)}}{\tau}\right)\\
&=\hspace{-2pt}\sum_{R_{m} \in \mathcal{M}_{i}} \hspace{-5pt}\tau \log \left( 1 + \exp \left( \sum_{P_{m'} \in B}\hspace{-5pt}\frac{S_{i, j, m, m'}^{(0)}}{\tau}\right)\!\right)\hspace{-5pt}\\
&= \tau \log \hspace{-5pt}\prod_{R_{m} \in \mathcal{M}_{i}}\hspace{-5pt} \left( 1 + \exp \left( \sum_{P_{m'} \in B}\hspace{-5pt}\frac{S_{i, j, m, m'}^{(0)}}{\tau}\right)\!\right)\\
&= \tau \log \sum_{A \subseteq \mathcal{M}_{i}} \prod_{R_{m} \in A} \exp \left( \sum_{P_{m'} \in B} \frac{S_{i, j, m, m'}^{(0)}}{\tau}\right)\\
&= \tau \log \sum_{A \subseteq \mathcal{M}_{i}} \exp \left( \sum_{R_{m} \in A} \sum_{P_{m'} \in B} \frac{S_{i, j, m, m'}^{(0)}}{\tau}\right)\\
&= \tau \log \sum_{A \subseteq \mathcal{M}_{i}} \exp \left( \frac{Q_{i, j, A, B}}{\tau}\right)
\end{align}
Then, from Lemma A.1, we have
\begin{align}
\hspace{-10pt}
\left| S^{(2)}_{i,j|B}\hspace{-2pt}-\!Q^{\leftarrow}_{i,j,B} \right|
&\!=\!
\left| 
\tau \log\hspace{-5pt}\sum_{A \subseteq \mathcal{M}_{i}}\hspace{-5pt}\exp\!\left(\!\frac{Q_{i,j,A,B}}{\tau}\!\right)
\!-\!Q^{\leftarrow}_{i,j,B} \right|\hspace{-5pt}\\
&\leq \tau M \log 2 .
\end{align}
Hence, for any $\epsilon>0$, choosing $\tau < \epsilon/(M \log 2)$
ensures
$\big| S^{(3)}_{i,j} - {Q}^{\leftarrow}_{i,j} \big| < \epsilon$.
Equivalently, $S^{(3)}_{i,j} \to {Q}^{\leftarrow}_{i,j}$ holds as $\tau \to 0^{+}$.
This completes the proof.
\end{proof}

\section*{Appendix B. Proof of Theorem 2}

In this section, we present a proof of Theorem 2. We first restate the definitions of the R2T aggregation and NLA-T2.

\subsection*{B.1 Preliminary}

\paragraph{R2T Aggregation}
We use the notation in Appendix A.1.
Given similarity $Q_{i,j,A,B}$, we define the R2T similarity as
\begin{align}
Q^{\rightarrow}_{i,j} := \frac{1}{2^{M}} \sum_{A\subseteq\mathcal{M}_i}
O_{i,j,A},
\end{align}
where $O_{i,j,A}$ evaluates the best-matching node $B$ for each region subset $A$ as
\begin{align}
O_{i,j,A} = \max_{B\in\mathcal{T}_j} Q_{i,j,A,B}.
\end{align}

\paragraph{Exponential Aggregation.}
Given similarity $Q_{i,j,A,B}$, we define exponential aggregation  $E_{i,j,B}$ as the sum of exponential similarities over the powerset of $\mathcal{M}_{i}$ with a temperature $\tau > 0$, \textit{i.e.},
\begin{align}
E_{i,j,B} = 
\sum_{A \subseteq \mathcal{M}_{i}}\exp\!\left( \frac{Q_{i,j,A,B}}{\tau}\right).
\end{align}

\paragraph{Bounding functions.}
For convenience, we define four auxiliary functions to evaluate upper and lower bounds:
\begin{align}
\Gamma_{B}(\alpha)&:=\frac{1-\alpha}{2}\,\bar{Q}_{i,j,B}+\alpha\max_{A} Q_{i,j,A,B}\\
\bar{\Gamma}_{B}(\tau,\alpha)
&:=
\frac{1-\alpha}{2}\,\bar{Q}_{i,j,B}+\alpha\,\tau\log E_{i,j,B}\\
\Lambda(\alpha)
&:= \max_{B \in \mathcal{T}_{j}} \Gamma_{B}(\alpha)\\
\bar{\Lambda}(\tau,\alpha)
&:= \tau \log \left( \sum_{B \in \mathcal{T}_{j}} \exp \left( \frac{\bar{\Gamma}_{B}(\tau,\alpha)}{\tau} \right) \right)
\end{align}

\paragraph{\textit{Lemma B.1 (Summation Over Powerset).}}
\textit{For any similarity $Q_{i,j,A,B}$, we have}
\begin{align}
E_{i,j,B} = 2^{M} \exp\!\left(\frac{\bar{Q}_{i,j,B}}{2\tau} \right)\!\hspace{-2pt}\prod_{R_{m} \in \mathcal{M}_{i}}
\hspace{-5pt}\cosh\left( \frac{Q_{i,j,m,B}}{2\tau} \right).\hspace{-5pt}
\end{align}
\begin{proof}
For any $x_{m} \in \mathbb{R}\;(m=1,2,\cdots M)$, we have
\begin{align}
\cosh(x_{m}) = \frac{1}{2}\left(\exp(x_{m}) + \exp(-x_{m})\right)
\end{align}
and thus
\begin{align}
\hspace{-7pt}
\prod_{m=1}^{M} \cosh(x_{m})
&=
\frac{1}{2^{M}}
\prod_{m=1}^{M} \left(\exp(x_{m}) + \exp(-x_{m})\right)\\
&=
\frac{1}{2^{M}}
\hspace{-5pt}\sum_{A \subseteq [M]}\hspace{-5pt}
\left(\prod_{m \in A}\hspace{-2pt}\exp(x_{m})\hspace{-2pt}\prod_{m \not \in A}\hspace{-2pt}\exp(-x_{m})\hspace{-2pt}\right)\\
&=
\frac{1}{2^{M}}
\hspace{-5pt}\sum_{A \subseteq [M]}\hspace{-5pt}
\exp \left(\sum_{m=1}^{M} (2\chi_{A}(m)-1) x_{m}\right)\\
&=
\frac{1}{2^{M}}
\exp(- \bar{x})
\hspace{-5pt}\sum_{A \subseteq [M]}\hspace{-5pt}
\exp \left(2 \sum_{m \in A} x_{m} \right)
\end{align}
where
\begin{align}
\chi_{A}(m)
&=
\begin{cases}
1 & (m \in A)\\
0 & (\text{otherwise})
\end{cases}
\\
\bar{x} &= \sum_{m=1}^{M} x_{m}
\end{align}
By substituting $x_{m} = Q_{i,j,m,B}/(2\tau)$,
 we obtain
\begin{align}
\hspace{-5pt}
\prod_{R_{m}\in\mathcal{M}_{i}}& \cosh\left(\frac{Q_{i,j,m,B}}{2\tau}\right)
=\nonumber\\
&\frac{1}{2^{M}}
\exp\left(- \frac{\bar{Q}_{i,j,B}}{2\tau}\right)
\hspace{-5pt}\underbrace{\sum_{A \subseteq \mathcal{M}_{i}}\hspace{-5pt}
\exp \left(\frac{Q_{i,j,A,B}}{\tau} \right)}_{E_{i,j,B}}.
\end{align}
Therefore, we have
\begin{align}
E_{i,j,B} = 2^{M} \exp\!\left(\frac{\bar{Q}_{i,j,B}}{2\tau} \right)\!\hspace{-2pt}\prod_{R_{m} \in \mathcal{M}_{i}}
\hspace{-5pt}\cosh\left( \frac{Q_{i,j,m,B}}{2\tau} \right)\hspace{-5pt}.
\end{align}
This completes the proof.
\end{proof}

\paragraph{Lemma B.2 (LSE Bound)}
For any $\tau > 0$ and $\alpha \in [0,1]$, we have
\begin{align}
\Lambda(\alpha)
\leq 
\bar{\Lambda} (\tau, \alpha)
\leq 
\Lambda(\alpha) + \tau Z_{j}
\end{align}
where $Z_{j} = \alpha M\log 2 + \log K_{j}$.

\begin{proof}
By the LSE inequality, for any $\tau > 0$ we have
\begin{align}
\max_{A} Q_{i,j,A,B}
\le \tau \log E_{i,j,B}
\le \max_{A} Q_{i,j,A,B} + \tau \log 2^{M}.
\end{align}
Multiplying by $\alpha \in [0,1]$ and adding $\tfrac{1-\alpha}{2}\,\bar{Q}_{i,j,B}$ to all terms gives
\begin{align}
\Gamma_{B}(\alpha)
\;\le\; \bar{\Gamma}_{B}(\tau,\alpha)
\;\le\; \Gamma_{B}(\alpha) + \tau \alpha M \log 2.
\label{eq:gamma_bound}
\end{align}
Next, applying the LSE inequality over $B$ to $\bar{\Lambda}(\tau,\alpha)$, we obtain
\begin{align}
\max_{B} \bar{\Gamma}_{B}(\tau,\alpha)
\;\le\; \bar{\Lambda}(\tau,\alpha)
\;\le\; \max_{B} \bar{\Gamma}_{B}(\tau,\alpha) + \tau \log K_{j}.
\end{align}
Combining this with Eq.~\eqref{eq:gamma_bound}, we obtain
\begin{align}
\Lambda(\alpha)
\;\le\; \bar{\Lambda}(\tau,\alpha)
\;\le\; \Lambda(\alpha) + \tau \alpha M \log 2 + \tau \log K_{j}.
\end{align}
This completes the proof.
\end{proof}

\subsection*{B.2 NLA-T2}

We define NLA-T2 and provide a proof for Theorem 2.

\paragraph{\textit{Definition 2 (NLA-T2).}}
\textit{NLA-T2 is a class of NLAs defined by the following activation functions and hyperparameters:}
\begin{align}
\hspace{-10pt}
\sigma_{1}(x)\!=\!\zeta_{\alpha}\!\left(\frac{x}{2\tau}\right),
\sigma_{2}(x) = \exp(x),\;
\sigma_{3}(x)\!=\!\tau \log(x),
\hspace{-10pt}
\end{align}
\textit{where $\zeta_{\alpha}(x) = x + \alpha \int \mathrm{Act}(x) \mathrm{d}x$ is a residual antiderivative of a differentiable activation function $\mathrm{Act}$, satisfying $\zeta_{\alpha}(0)=0$, and $\tau$ is a temperature hyperparameter.}

\begin{figure*}
\centering
\includegraphics[width=\linewidth]{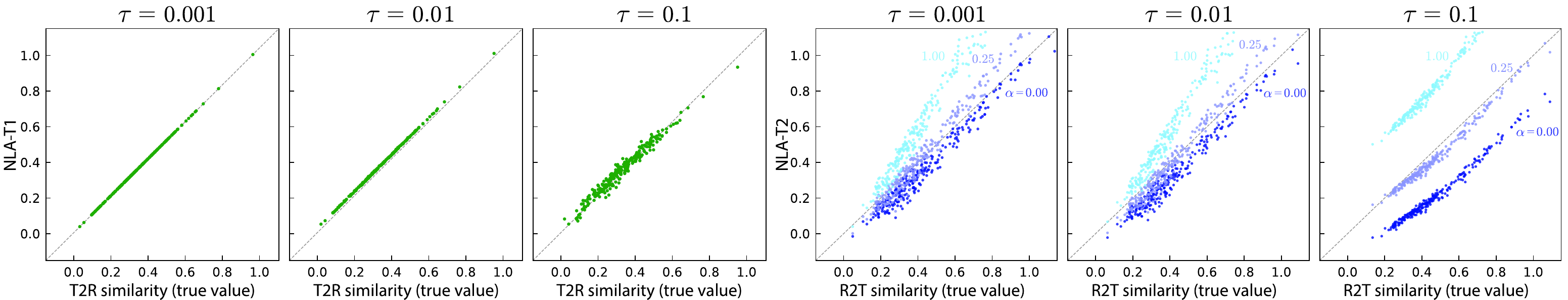}
\caption{Approximation accuracy evaluation for NLA-T1 and NLA-T2.}
\label{fig:fig_powerclip_app_nla}
\end{figure*}

\begin{table*}[!t]
\centering
\scriptsize
\midsepremove
\setlength{\tabcolsep}{2pt}
\renewcommand{\arraystretch}{1.2}

\begin{tabular}{l|*{17}{c}}
\toprule
\multirow{2}{*}{Method} &
\multicolumn{17}{c}{Zero-shot classification (Top-1)} \\
\cmidrule(lr){2-18}
& Food101 & CIFAR10 & CIFAR100 & SUN397 & Cars & VOC07 & Aircraft & DTD & Pets & Cal101 & Flowers & STL10 & EuroSAT & RESISC45 & GTSRB & Country & PCam \\
\midrule
PreviousSOTA
& 46.5      & 74.3      & 37.3      & 47.5      & 19.9      & \textbf{61.1} & \textbf{3.1} & 19.8      & 58.1      & \underline{73.7} & 30.7      & \textbf{92.8} & \textbf{33.2} & 30.4      & \textbf{10.9} & \underline{4.8}  & \underline{50.8} \\
\rowcolor{lightcyan}\textbf{PowerCLIP-R}
& \underline{50.3} & \underline{74.7} & \textbf{43.5} & \underline{48.7} & \underline{22.9} & 53.2      & \underline{2.9} & \textbf{21.5} & \underline{58.7} & \textbf{75.7} & \underline{32.4} & 88.4      & \underline{30.8} & \textbf{37.5} & \underline{9.8}  & 4.6         & 50.0      \\
\rowcolor{lightcyan}\textbf{PowerCLIP-S}
& \textbf{51.2} & \textbf{81.3} & \underline{40.1} & \textbf{50.5} & \textbf{23.5} & \underline{56.0} & 1.6       & \underline{21.3} & \textbf{61.0} & 72.9      & \textbf{32.5} & \underline{90.5} & 29.0      & \underline{33.9} & 7.8       & \textbf{5.4}  & \textbf{59.7} \\
\bottomrule
\end{tabular}

\vspace{3pt}

\begin{tabular}{l|
*{6}{c}|
*{6}{c}|
*{3}{c}|
*{2}{c}}
\toprule
\multirow{2}{*}{Method} &
\multicolumn{6}{c|}{Zero-shot retrieval R@1} &
\multicolumn{6}{c|}{Robustness (Top-1)} &
\multicolumn{3}{c|}{SugarCrepe} &
\multicolumn{2}{c}{Winoground} \\
\cmidrule(lr){2-7}
\cmidrule(lr){8-13}
\cmidrule(lr){14-16}
\cmidrule(lr){17-18}
& COCO-T & F8K-T & F30K-T & COCO-I & F8K-I & F30K-I
& IN-1k & IN-V2 & IN-A & IN-R & IN-O & IN-S
& Obj & Att & Rel
& Text & Image \\
\midrule
PreviousSOTA
& 36.0 & 58.3 & 59.9 & 25.1 & 44.4 & \underline{47.1}
& 38.6 & 33.1 & 9.6 & 48.1 & \textbf{42.6} & 25.6
& 75.5 & \textbf{70.8} & \textbf{67.9}
& \textbf{25.2} & \underline{13.5} \\
\rowcolor{lightcyan}\textbf{PowerCLIP-R}
& \underline{36.7} & \underline{58.5} & \underline{61.7} & \underline{26.3} & \underline{44.8} & 46.6
& \underline{40.3} & \underline{34.8} & \underline{11.2} & \underline{53.2} & 40.2 & \underline{28.7}
& \underline{75.6} & 70.3 & \textbf{67.9}
& 22.5 & 9.5 \\
\rowcolor{lightcyan}\textbf{PowerCLIP-S}
& \textbf{37.3} & \textbf{58.6} & \textbf{62.4} & \textbf{27.0} & \textbf{46.3} & \textbf{50.4}
& \textbf{40.8} & \textbf{35.1} & \textbf{11.9} & \textbf{53.5} & \underline{40.5} & \textbf{28.9}
& \textbf{76.1} & \underline{70.4} & 67.1
& \underline{24.8} & \textbf{16.0} \\
\bottomrule
\end{tabular}

\caption{
Detailed unified comparison of PreviousSOTA (best over CLIP–SPARC), PowerCLIP-R, and PowerCLIP-S.
Top: 17 zero-shot classification datasets.
Bottom: 6 zero-shot retrieval settings (R@1), 6 ImageNet robustness benchmarks, SugarCrepe compositionality, and Winoground compositionality.
}
\label{tab:detailed_unified_powerclip}
\end{table*}

\vspace{2pt}
\paragraph{\textit{Theorem 2.}} \textit{Suppose $\mathrm{Act}=\tanh$. Then, NLA-T2 approximates the R2T similarity ${Q}^{\rightarrow}_{i,j}$ with arbitrary precision.
That is, for any $\epsilon>0$, there exist $\tau>0$ and $\alpha\in[0,1]$ such that $\big|S^{(3)}_{i,j}-{Q}^{\rightarrow}_{i,j}\big|<\epsilon$.}

\begin{proof}
With $\mathrm{Act}=\tanh$, we have
\begin{align}
\int \tanh(x)\,\mathrm{d}x = \log \cosh(x)    
\end{align}
and thus, with $\zeta_{\alpha}(0)=0$, we obtain
\begin{align}
\zeta_{\alpha}(x)=x+\alpha \log\!\cosh(x).
\end{align}
Then, the output of the first layer is given by
\begin{align}
S^{(1)}_{i,j,m| B}
&=\zeta_{\alpha}\!\left(\frac{1}{2\tau}\sum_{P_{m'}\in B} S^{(0)}_{i,j,m,m'}\right)\\
&= \frac{Q_{i,j,m,B}}{2\tau}+\alpha \log\!\cosh\!\left(\frac{Q_{i,j,m,B}}{2\tau}\right).
\end{align}
Applying $\sigma_{2}(x)\!=\!\exp(x)$ at the second layer, we have,
\begin{align}
S^{(2)}_{i,j| B}
&=\exp\!\left(\sum_{R_{m} \in \mathcal{M}_{i}} S^{(1)}_{i,j,m| B}\right)\\
&=\exp\!\left(\frac{\bar{Q}_{i,j,B}}{2\tau}\right) \prod_{R_{m} \in \mathcal{M}_{i}}\hspace{-5pt}\cosh^{\alpha}\!\left(\frac{Q_{i,j,m,B}}{2\tau}\right).
\end{align}
From Lemma B.1, we have
\begin{align}
S^{(2)}_{i,j| B}
&=2^{-\alpha M}\exp\!\left(\frac{1-\alpha}{2\tau}\,\bar{Q}_{i,j,B}\right)\left(E_{i,j,B}\right)^{\alpha}\\
&=2^{-\alpha M}\exp\!\left(\frac{\bar{\Gamma}_{B}(\tau,\alpha)}{\tau}\right),
\end{align}
Applying $\displaystyle \sigma_{3}(x)\!=\!\tau\log(x)$ at the third layer, we have
\begin{align}
S^{(3)}_{i,j}
=&\tau\log\!\left(\frac{1}{|\mathcal{T}_{j}|^{\,1-\alpha}}\sum_{B\in\mathcal{T}_{j}} S^{(2)}_{i,j| B}\right) \\
=& -\tau Z^{\alpha}_{j} + \tau\log\!\left(\sum_{B\in\mathcal{T}_{j}}\exp\!\left(\frac{\bar{\Gamma}_{B}(\tau,\alpha)}{\tau}\right)\right)\\
=& -\tau Z^{\alpha}_{j} + \bar{\Lambda}(\tau, \alpha),
\end{align}
where $Z^{\alpha}_{j}=\alpha M \log 2 + (1-\alpha) \log K_{j}$.
From Lemma B.2, we obtain the quantitative bounds
\begin{align}
\Lambda(\alpha) - \tau Z^{\alpha}_{j}
\;\le\;
S^{(3)}_{i,j}
\;\le\;
\Lambda(\alpha)+\tau \alpha \log K_{j}.
\end{align}
In particular,
\begin{align}
\lim_{\tau\to 0^{+}} S^{(3)}_{i,j}
=\Lambda(\alpha).
\end{align}
Since $\mathbb{E}_{A}\!\left[Q_{i,j,A,B}\right]=\tfrac{1}{2}\bar{Q}_{i,j,B}$ under the uniform distribution over $2^{\mathcal{M}_{i}}$, Jensen's inequality for the pointwise maximum implies
\begin{align}
\Lambda(0) = \frac{1}{2}\max_{B}\bar{Q}_{i,j,B}
\;\le\;
Q^{\rightarrow}_{i,j}
\;\le\;
\max_{A,B} Q_{i,j,A,B}
=\Lambda(1).
\end{align}
Because $\Lambda(\alpha)$ is continuous in $\alpha\in[0,1]$ (being the maximum of finitely many affine functions of $\alpha$), there exists $\alpha^{\star}\in[0,1]$ such that $\Lambda(\alpha^{\star})=Q^{\rightarrow}_{i,j}$.

Finally, by the quantitative bounds above, we obtain
\begin{align}
\left|S^{(3)}_{i,j}-Q^{\rightarrow}_{i,j}\right|
&=\left|S^{(3)}_{i,j}-\Lambda(\alpha^{\star})\right|\\
&\le \tau\!\left(\alpha^{\star} M\log 2+\log K_{j}\right).
\end{align}
Hence, for any $\epsilon{>}0$, choosing $\tau{<}\epsilon\big/\!\left(\alpha^{\star} M\log 2+\log K_{j}\right)$ ensures
$\left|S^{(3)}_{i,j}-Q^{\rightarrow}_{i,j}\right|<\epsilon$.
This shows that NLA-T2 approximates ${Q}^{\rightarrow}_{i,j}$ with arbitrary precision, completing the proof.
\end{proof}

\section*{Appendix C. Approximation Accuracy}
To quantitatively evaluate the approximation accuracy of NLAs, we compare the approximated similarity values with the true values.
Figure~\ref{fig:fig_powerclip_app_nla} shows the outputs from NLA-T1 and NLA-T2 compared against the true T2R and R2T similarity values computed on synthetic data (randomly generated input vectors) with all parameters initialized randomly.
For NLA-T1, approximations closely correlate with the true values.
For NLA-T2, although the distribution displays greater variance compared to NLA-T1, the correlation remains strong.
Additionally, we see that NLA-T2 with $\alpha\!=\!1.0$ and $\alpha\!=\!0.0$ corresponds to the upper and lower bounds of the R2T similarity, respectively.
However, when $\tau = 0.1$, approximations are biased, resulting in larger errors during loss computation.
These results are consistent with our theoretical analysis and validate the effectiveness of the proposed NLAs.

\section*{Appendix D. Details of Evaluation Task}
For a comprehensive comparison with prior methods, Table~\ref{tab:detailed_unified_powerclip} provides the tabular counterpart of the performance summary shown in Figure~\ref{fig:performance}. The table reports results on 17 zero-shot classification datasets, 6 zero-shot retrieval R@1 settings, 6 ImageNet-based robustness benchmarks, as well as compositional generalization performance on SugarCrepe and Winoground under a unified evaluation protocol.
\begin{table}[ht]
\centering
\midsepremove
\small
\setlength{\tabcolsep}{4.1pt}
\begin{tabular}{l|ccccccc}
\toprule
\multirow{2}{*}{Method} 
& \multicolumn{3}{c}{SC-REPLACE} & \multicolumn{2}{c}{SC-SWAP} & \multicolumn{2}{c}{SC-ADD} \\
\cmidrule(lr){2-4} \cmidrule(lr){5-6} \cmidrule(lr){7-8}
& Obj & Att & Rel & Obj & Att & Obj & Att \\
\midrule
CLIP~\cite{clip} & 85.8 & \textbf{79.2} & 64.5 & \underline{61.8} & 58.7 & 74.2 & 68.4 \\
FLIP~\cite{Li2023flip} & 84.1 & 75.9 & 66.0 & 60.2 & 61.6 & 71.7 & 63.2 \\
A-CLIP~\cite{Yang2023a-clip} & 86.6 & 75.5 & 63.2 & 52.4 & 63.1 & 71.6 & 66.8 \\
E-CLIP~\cite{Wei2024e-clip} & 86.9 & 73.5 & 60.2 & 59.4 & 63.4 & 73.3 & 66.8 \\
C-PGS~\cite{Pei2025CLIPPGS} & \underline{88.1} & 76.0 & \textbf{67.9} & \textbf{64.1} & \underline{66.5} & 74.2 & \underline{69.9} \\
FILIP~\cite{Yao2022FILIP} & 82.9 & 61.9 & 56.8 & 58.4 & 58.3 & 53.4 & 54.3 \\
SPARC~\cite{Bica2024SPARC} & 85.2 & 75.5 & 66.9 & 58.8 & \textbf{67.4} & 76.4 & 68.6 \\
\rowcolor{lightcyan} PowerCLIP-R & \textbf{88.3} & 76.6 & \underline{67.8} & 60.8 & 64.7 & \underline{77.8} & 69.7 \\
\rowcolor{lightcyan} PowerCLIP-S & 87.5 & \underline{77.5} & 67.1 & 61.6 & 63.1 & \textbf{79.1} & \textbf{70.7} \\
\bottomrule
\end{tabular}
\vspace{-0.2cm}
\caption{
Detailed compositionality evaluation on SugarCrepe.
}
\label{app:tab:results_compositionality}
\end{table}
\begin{figure*}[t]
\centering
\begin{minipage}{0.32\textwidth}
    \centering
    \includegraphics[width=\linewidth]{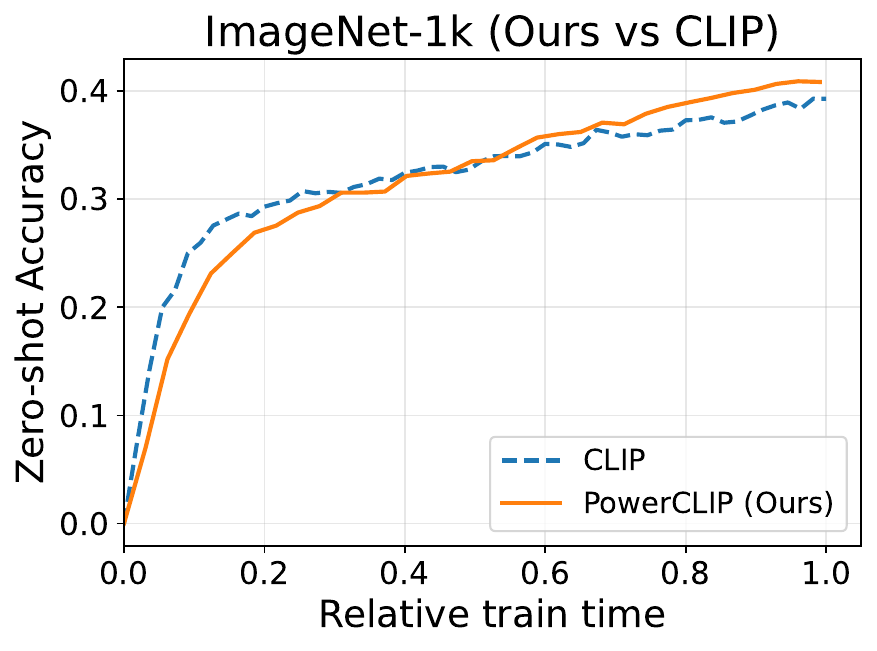}
    \captionof{figure}{\textbf{ImageNet-1k zero-shot accuracy vs relative training time.} CLIP is trained for more epochs so that its total compute matches our method.}
    \label{fig:app_relative_train}
\end{minipage}
\hfill
\begin{minipage}{0.32\textwidth}
    \centering
    \includegraphics[width=\linewidth]{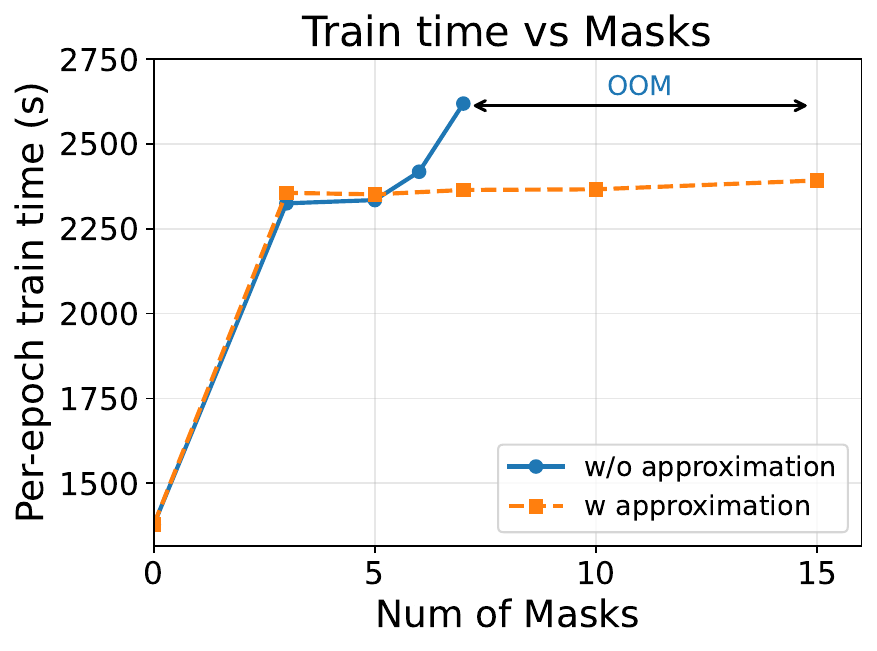}
    \captionof{figure}{\textbf{Per-epoch training time vs number of masks $K$ with and without approximation.} Without approximation, runs with $K{>}7$ fail due to OOM.}
    \label{fig:time_vs_topk}
\end{minipage}
\hfill
\begin{minipage}{0.32\textwidth}
    \centering
    \small
    \midsepremove
    \setlength{\tabcolsep}{4pt}
    \begin{tabular}{l|cc}
    \toprule
    Method & Train time (s) & Rel. to CLIP \\
    \midrule
    CLIP~\cite{clip} & 1378 & $1.00\times$ \\
    SPARC~\cite{Bica2024SPARC} & 1730 & $1.26\times$ \\
    FILIP~\cite{Yao2022FILIP} & 1947 & $1.41\times$ \\
    PowerCLIP & 2366 & $1.72\times$ \\
    \bottomrule
    \end{tabular}
    \captionof{table}{\textbf{Per-epoch training time of each method under our training setup.} We report wall-clock time in seconds and relative cost normalized by CLIP.}
    \label{tab:appendix_train_time}
\end{minipage}
\end{figure*}
For SugarCrepe, Table~\ref{tab:results_compositionality} in the main paper reports only the average scores for each method, whereas Table~\ref{app:tab:results_compositionality} presents a more fine-grained breakdown. In particular, our method tends to show better performance than prior approaches in the replace and add settings, suggesting that it more effectively improves text–image consistency under more challenging compositional transformations.

\section*{Appendix E. Computational Cost}
\label{sec:appendix_cost}
Table~\ref{tab:appendix_train_time} reports the per-epoch training time compared to prior work, showing that our method incurs about a $1.72\times$ higher cost than CLIP due to the additional computation from region-level features and parse-tree reasoning. To account for this overhead and compare under a matched compute budget, we train CLIP for roughly $1.72\times$ more epochs (from 32 to about 55 epochs) and evaluate ImageNet-1k zero-shot performance. Figure~\ref{fig:app_relative_train} shows the resulting accuracy as a function of relative training time, demonstrating that even under the same total training cost, our method still outperforms CLIP.
Figure~\ref{fig:time_vs_topk} shows how the per-epoch training time changes with the number of masks, with and without our approximation. Without approximation, the training time already starts to grow noticeably at 6 masks, and increasing the number of masks beyond 7 leads to out-of-memory (OOM) failures. In contrast, with our approximation, we can safely scale the number of masks up to 15 while keeping the per-epoch training time only mildly increased. This demonstrates that the proposed approximation effectively reduces both computation and memory overhead, enabling the use of richer region-level information within a practical training budget.

\section*{Appendix F. Ablation Study on $\lambda$}
\label{sec:appendix_lambda}

We study the sensitivity of PowerCLIP to the mixing coefficient $\lambda$ under the main training setting in Sec.~\ref{sec:powerset_alignment}.
Table~\ref{tab:lambda_ablation} reports classification (Cls) and retrieval (Ret) performance for $\lambda \in \{0.1, 0.2, 0.3\}$.
While retrieval is relatively stable, classification varies more across $\lambda$.
We therefore use $\lambda=0.1$ in all experiments as it yields a balanced trade-off between retrieval and classification.

\section*{Appendix G. Open-Vocabulary Evaluation on OV-COCO}
\label{sec:appendix_ov_coco}

To further examine fine-grained recognition beyond closed-set benchmarks, we evaluate PowerCLIP on OV-COCO~\cite{zareian2021open}.
As shown in Table~\ref{tab:ov_detection}, PowerCLIP improves over CLIP by 7.2 points in $\mathrm{AP}_{50}$ and by 13.9 points in $\mathrm{AP}^{\mathrm{novel}}_{50}$.
These gains indicate that PowerCLIP captures finer-grained visual--text correspondences, particularly for novel categories.

\begin{table}[t]
\centering
\small
\setlength{\tabcolsep}{4pt}

\begin{minipage}[t]{0.38\linewidth}
  \centering
  \setlength{\tabcolsep}{4pt}
  \midsepremove
  \begin{tabular}{lcc}
    \toprule
    Value & Cls & Ret\\
    \midrule
    $\lambda\!=\!0.1$ & 42.2 & 47.0\\
    $\lambda\!=\!0.2$ & 40.7 & 47.6\\
    $\lambda\!=\!0.3$ & 41.3 & 47.1\\
    \bottomrule
  \end{tabular}
  \captionof{table}{Ablation for $\lambda$.}
  \label{tab:lambda_ablation}
\end{minipage}\hfill
\begin{minipage}[t]{0.6\linewidth}
  \centering
  \setlength{\tabcolsep}{2pt}
  \midsepremove
  \begin{tabular}{lccc}
    \toprule
    Method & $\mathrm{AP}^{\mathrm{base}}_{50}$ & $\mathrm{AP}^{\mathrm{novel}}_{50}$ & $\mathrm{AP}_{50}$ \\[-0.1em]
    \midrule
    CLIP~\cite{clip}    & 22.8 & 1.4 & 17.2 \\
    FLIP~\cite{Li2023flip}    & 24.1 & 0.9 & 18.0 \\
    FILIP~\cite{Yao2022FILIP}   & 21.6 & 3.2 & 16.8 \\
    SPARC~\cite{Bica2024SPARC}   & 20.8 & 7.1 & 17.2 \\
    C-PGS~\cite{Pei2025CLIPPGS}   & 23.1 & 2.3 & 17.7 \\
    \rowcolor{lightcyan} PowerCLIP & 27.6 & 15.3 & 24.4 \\
    \bottomrule
  \end{tabular}
  \captionof{table}{Evaluation on OV-COCO.}
  \label{tab:ov_detection}
\end{minipage}

\vspace{-12pt}
\end{table}

\section*{Appendix H. Qualitative Examples}
Figure~\ref{fig:app_qualitative_compare} shows a side-by-side comparison of text–image patch similarity heatmaps with existing models for the same inputs as in Figure~\ref{fig:qualitative}. Compared to prior methods, our model produces sharper and more localized activations for words, indicating a closer alignment between the textual structure and the corresponding image regions.
Figure~\ref{fig:app_qualitative_compositional} illustrates the compositional reasoning ability of our model by intentionally altering the order and attributes of objects in the text. When we apply such compositional edits to the caption, the high-similarity regions in the image shift accordingly to the appropriate patches, demonstrating that our model maintains semantically consistent text–image correspondences under such compositional transformations.

\begin{figure*}
\centering
\includegraphics[width=\linewidth,height=9cm]{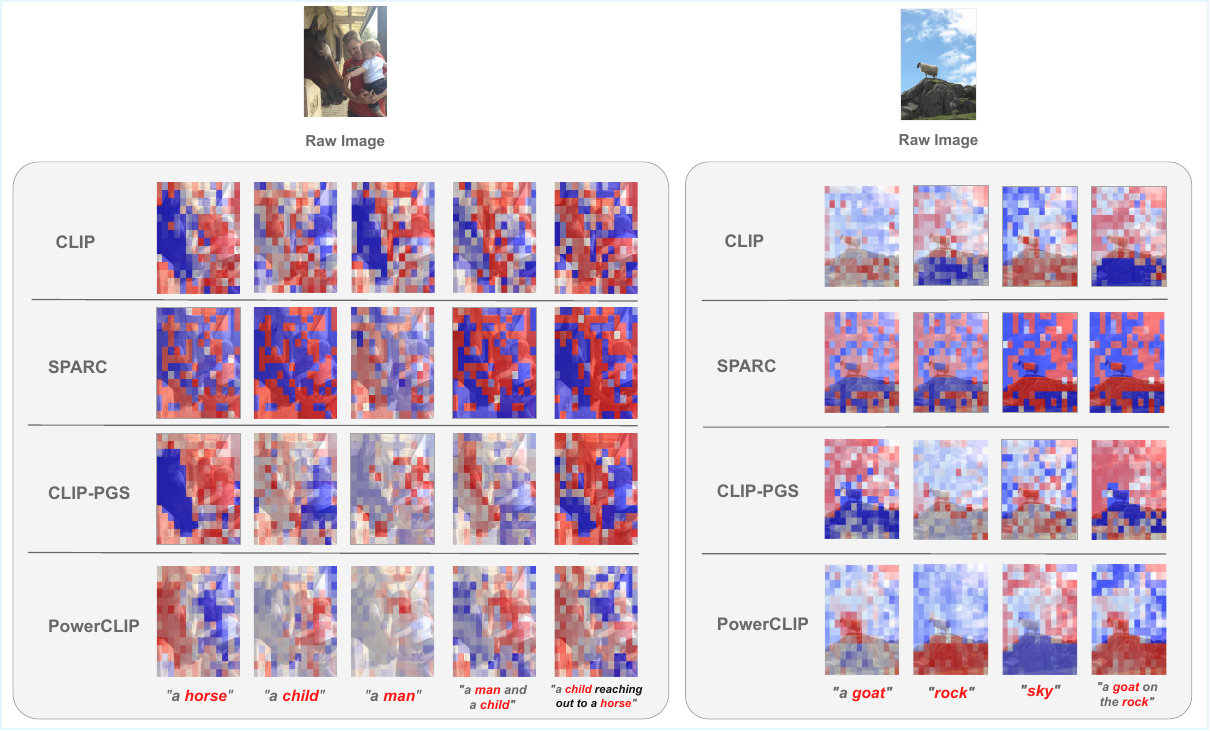}
\captionof{figure}{Qualitative comparison of text-to-patch similarity heatmaps across different models.}
\label{fig:app_qualitative_compare}
\end{figure*}
\begin{figure*}
\centering
\includegraphics[width=\linewidth]{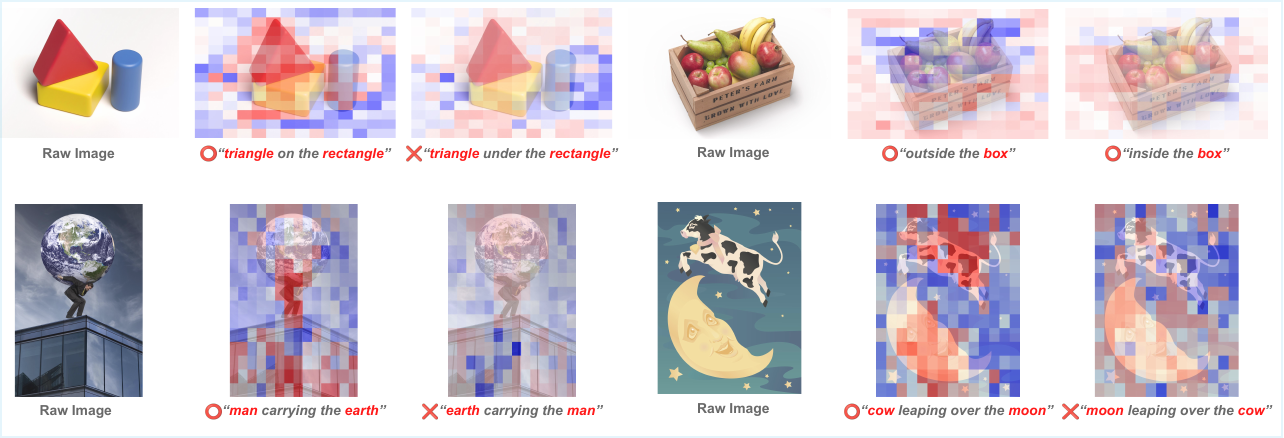}
\captionof{figure}{Qualitative examples of compositional reasoning.}
\label{fig:app_qualitative_compositional}
\end{figure*}

\end{toggle}

\end{document}